\documentclass{article}
\usepackage{iclr2027_conference,times}
\usepackage{microtype}
\usepackage{graphicx}
\usepackage{wrapfig}
\usepackage{subcaption}
\usepackage{booktabs}
\usepackage{diagbox}
\usepackage{multirow}
\usepackage{tabularx}
\usepackage{amsmath}
\usepackage{amssymb}
\usepackage{mathtools}
\usepackage{amsthm}
\usepackage{pifont}
\usepackage{float}
\usepackage{algorithm}
\usepackage{algorithmic}
\usepackage{hyperref}
\usepackage{url}
\usepackage[capitalize,noabbrev]{cleveref}

\theoremstyle{plain}

\theoremstyle{definition}

\theoremstyle{remark}

\newcommand{\paperfigure}[2]{%
  \IfFileExists{#2}{\includegraphics[width=#1]{#2}}{%
    \PackageWarning{ELV}{Missing figure: #2. Restore the original asset before submission}%
    \fbox{\parbox[c][0.16\textheight][c]{\dimexpr#1-2\fboxsep-2\fboxrule\relax}{%
      \centering\small\textbf{Missing figure}\\\nolinkurl{#2}\\Original asset required.}}%
  }%
}
\iclrfinalcopy

\author{Yijie Sun$^1$, Sanquan Sun$^2$, Yanda Zhu$^1$, Yuanyang Zhu$^1$,
\textbf{Yaohua Hu}$^3$, \textbf{Chunlin Chen}$^{1}$
\thanks{Correspondence to Chunlin Chen.}
\vspace{0.5em}\\
$^1$ Nanjing University~~~~~~~~$^2$ Hong Kong Polytechnic University~~~~~~~~ 
$^3$ Shenzhen University
\vspace{0.5em}\\
\texttt{2796624556qw@gmail.com}~~~~~~~~~~~~~~~~~~~~~~~~~\texttt{sanquan.sun@connect.polyu.hk} \\
\texttt{yandazhu@smail.nju.edu.cn}~~~~~~~~~~~~~~~~~~~\texttt{mayhhu@szu.edu.cn} \\
\texttt{yuanyangzhu@nju.edu.cn}~~~~~~~~~~~~~~~~~~~~~~~~~~\texttt{clchen@nju.edu.cn} \\
}
\title{Escaping Local Views: Discovering Latent Concepts for Interpretable Multi-Agent Reinforcement Learning}

\begin{document}
\maketitle
\lhead{Preprint}

\begin{abstract}
Efficient cooperation is challenging due to the usual partial observability of each agent in multi-agent reinforcement learning.
Recurrent networks encode local interaction histories, but their hidden representations provide limited insight into the information underlying individual decisions.
To address these challenges, we propose a novel interpretable framework, called escaping local views (ELV), which introduces semantically structured latent concepts to render policy decisions transparent.
Specifically, each agent extracts low-dimensional semantic concepts from its local observation and action-observation trajectory.
These concepts are jointly encoded into a contextual latent variable via a variational autoencoder (VAE), which builds a bridge between local views and global semantics.
To explicitly model the decision of each agent, we employ a dual-path attention mechanism in which one module estimates the salience of individual concepts relative to the global context, while the other captures higher-order cooperative patterns with pairwise concept interactions.
Furthermore, we incorporate a concept prediction module that derives an intrinsic reward from next-concept prediction errors, which incentivizes agents to explore regions of semantic novelty.
Experiments in multiple environments verify that ELV not only achieves competitive performance but also explicitly provides how agents reason about their decisions.
\end{abstract}

\section{Introduction}
Cooperative multi-agent reinforcement learning (MARL) has demonstrated strong empirical performance on a range of complex decision-making problems, from multi-robot coordination and large-scale fleet management to autonomous driving~\citep{cao2012overview, liu2024autodrive} and real-time strategy games~\citep{vinyals2019grandmaster, geng2025l2m2}. 
However, the policies learned by modern deep MARL algorithms are typically parameterized by high-capacity neural networks, whose inherently opaque representations pose significant barriers to interpretability and hinder adoption in high-stakes applications~\citep{boggess2023explainable}.
In addition, many cooperative MARL problems are naturally modeled as partially observable Markov decision processes (POMDPs), where each agent has access only to its own local observations while the environment's global state remains hidden.
To address the resulting non-stationarity and information asymmetry, the centralized training with decentralized execution (CTDE) paradigm has become the de facto standard: agents leverage centralized information during training but act solely based on local observations at execution time.
While CTDE mitigates some optimization challenges, the split between centralized training and decentralized execution can fragment local representations and coordination, degrading generalization and destabilizing learned policies~\citep{du2025multi,barde2024model}.

A natural response is to equip MARL with interpretability, which largely falls into two categories.
The first class provides \emph{post-hoc} explanations for already-trained policies, typically by constructing local approximations of the decision function, e.g., via Shapley-style attributions~\citep{wang2020shapley}, trajectory clustering~\citep{pmlr-v48-zahavy16}, or saliency-based visualizations~\citep{ghorbani2019interpretation}. 
While useful, such explanations are inherently approximate and could be computationally expensive~\citep{slack2021reliable}, and they may lead to incorrect explanations due to compounding error~\citep{ghorbani2019interpretation}, raising concerns about faithfulness and reliability.
A second line aims for \emph{intrinsically interpretable} policies by replacing a black-box agent with a simpler, human-readable surrogate (e.g., rule lists or decision trees), yielding a global summary of behavior. 
They generally explain decisions by inspecting inputs and outputs by visualizing intermediate weights~\citep{liu2025mixrts} or credit assignment~\citep{liu2023na2q,yang2020qatten,wang2022shaq} through value decomposition.
However, their explanations based on observation-level activations may not provide the most intuitive form of explanations for humans.

The quest for intrinsic explanations has led to a renewed interest in concept learning~\citep{koh2020concept,stammer2022interactive,havasi2022addressing}.
In particular, concept bottleneck models (CBMs)~\citep{koh2020concept} factorize prediction through an explicit concept layer: the model first infers a set of human-specified concepts and then maps these concepts to the final output, which yields a global, semantically grounded account of what drives decisions and enables test-time interventions by directly editing incorrect concepts~\citep{ramaswamy2023overlooked}.
Subsequent studies sharpen this view by (i) analyzing and mitigating concept leakage, where the network bypasses the bottleneck~\citep{kim2023probabilistic}, and (ii) formalizing interactive intervention mechanisms that improve controllability and reliability of concept edits~\citep{vandenhirtz2024stochastic}.
These advances motivate a natural question in MARL: can agents discover and share concept-level abstractions that support coordination under partial observability, while retaining the intervention and inspection benefits of CBMs?

We answer this question with \textbf{ELV}, a concept-bottleneck framework for cooperative MARL that addresses both interpretability and the local view limitation of CTDE. 
Each agent extracts a small set of latent concepts from its action--observation history, and uses these concepts as the sole input to value estimation. 
To align semantics across agents, ELV aggregates the distributed concept set and learns a \emph{variational semantic bridge} that reconstructs the global state during centralized training, thereby shaping the concept space to retain globally coherent information. 
The resulting latent context serves as a global query that reweights concepts through (i) \emph{first-order} contextual salience and (ii) \emph{second-order} interaction salience, so value estimation reflects not only what matters globally, but also what matters for coordination. 
Finally, ELV augments exploration with a concept-prediction intrinsic reward, which measures novelty in the learned concept dynamics and is less sensitive to task-irrelevant variation in raw observations.

Our contributions are: 1) We propose \textbf{ELV}, a plug-and-play concept-bottleneck module for CTDE MARL that computes individual value of each agent in a structured concept space, with explicit concept presence and a variational semantic bridge to align local concepts across agents.
2) We improve interpretability by \textbf{grounding concepts to local semantics} via an auxiliary \textbf{observation reconstruction} objective, enabling concept--feature attribution and semantic analysis.
3) We evaluate ELV on the Level-based foraging (LBF)~\citep{lbf}, StarCraft Multi-Agent Challenge (SMAC)~\citep{SMAC}, and SMACv2~\citep{ellis2023smacv2} benchmarks, where it achieves strong performance while learning coherent concepts that support interpretable coordination analysis.

\section{Preliminaries}
\textbf{Dec-POMDP.} A fully cooperative multi-agent task is typically formulated as a Decentralized Partially Observable Markov Decision Process (Dec-POMDP), consisting of a tuple $\langle\mathcal{N}, \mathcal{S}, \mathcal{U}, \mathcal{P}, r, O, \Omega, \gamma\rangle$, where $\mathcal{N}$ is a finite set of $n$ agents, $s \in S$ is the global state of the environment.
At each timestep $t$, every agent $i \in \mathcal{N}$ chooses an action $u_i \in \mathcal{U}$ to formulate a joint action $\boldsymbol{u}=\left[u_i\right]_{i=1}^n \in \mathcal{U}^n$ by its own observation $o_i \in \Omega$ received from the environment.
The environment returns a shared reward with the reward function $r(s, \boldsymbol{u}): \mathcal{S} \times \mathcal{U}^n \rightarrow \mathbb{R}$, and transits to the next state $s^{\prime}$ with the transition function $\mathcal{P}\left(s^{\prime} \mid s, \boldsymbol{u}\right): \mathcal{S} \times \mathcal{U}^n \rightarrow \mathcal{S}$.
The action-observation history of agent $i$ is denoted as $\tau_i \in \mathcal{T} \equiv(\Omega \times \mathcal{U})^*$ and the joint action-observation history is $\boldsymbol{\tau} \in \mathcal{T}^n$.
Agent $i$ learns its own policy $\pi_i\left(u_i \mid \tau_i\right): \mathcal{T} \times \mathcal{U} \rightarrow[0,1]$ conditions on its local action-observation history.
The objective in a Dec-POMDP is to find a joint policy $\boldsymbol{\pi}=\left\langle\pi_1, \ldots, \pi_n\right\rangle$, which maximizes the expected cumulative discounted reward.
The joint action-value function under a joint policy $\boldsymbol{\pi}$ is $Q^\pi(s, \boldsymbol{u})=r(s, \boldsymbol{u})+\gamma \mathbb{E}_{s^{\prime}}\left[R_t \mid s, \boldsymbol{u}\right]$, where $\gamma \in[0,1)$ is the discount factor and $R_t=\sum_{t=0}^{\infty} \gamma^t r_t$ is the cumulative return. 

\textbf{Concept Bottleneck Models (CBMs).} Concept bottleneck models (CBMs) introduce an explicit intermediate representation to separate concept extraction from prediction.
A concept extractor $g$ maps an input representation (e.g., a recurrent history state $h_i$) to a concept vector $c_i=g(h_i)$, and a predictor $f$ maps concepts to task outputs.
In our setting, this yields the action values of agent $i$
\begin{equation}
    Q(\tau_i, u_i) = f(g(h_i)).
\end{equation}
This factorization mediates decisions through structured variables rather than opaque latent features.

\textbf{Variational Autoencoders (VAEs).}
VAEs define a latent-variable model $p_\theta(s)=\int p_\theta(s\mid z)p(z)\,dz$, where $s\in\mathcal{S}$ denotes observed data and $z\in\mathbb{R}^d$ is a latent variable with prior $p(z)=\mathcal{N}(0,I)$.
Since the marginal likelihood $\log p_\theta(s)$ is intractable in general, VAEs introduce an approximate posterior $q_\phi(z\mid s)$ and maximize the evidence lower bound (ELBO):
\begin{equation}
\log p_\theta(x) \geq \mathbb{E}_{q_\phi(z \mid s)}\left[\log p_\theta(s\mid z)\right]- \mathrm{KL}\left(q_\phi(z \mid s) \| p(z)\right).
\end{equation}
It balances accurate reconstruction of $o$ with a regularizer that keeps $q_\phi(z\mid o)$ close to the prior.
Equivalently, the standard VAE objective is obtained by minimizing the negative ELBO:
\begin{equation}
\mathcal{L}_{\mathrm{VAE}}(s)=-\mathbb{E}_{q_\phi(z \mid s)}[\log p_\theta(s \mid z)]+ \mathrm{KL}(q_\phi(z \mid s) \| p(z)).
\end{equation}
The introduction of related works is presented in Appendix~\ref{RW}.

\begin{figure}[!t]
\centering
	\paperfigure{0.98\textwidth}{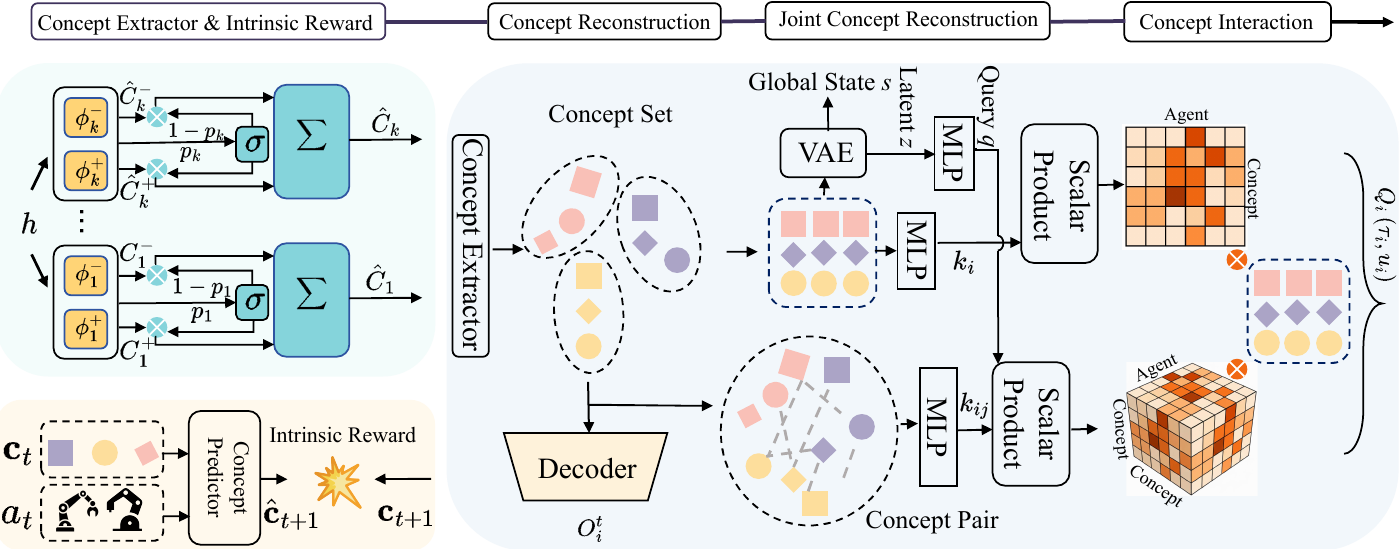}
	\caption{At each timestep $t$, agent $i$ encodes its local history $\tau_i^t$ into a hidden state $h_i^t$, from which ELV extracts concept embeddings $\{c_{i,k}^t\}_{k=1}^K$ and keys $\{\boldsymbol{k}_{i,k}\}$ via shared projections. 
The concepts of all agents are aggregated and passed through a VAE to infer a global context $\boldsymbol{z}$, which yields a query vector $\boldsymbol{q}$.
First-order weights $\alpha_{i,k}$ are obtained by softmax-normalizing dot-product scores $\boldsymbol{q}^\top \boldsymbol{k}_{i,k}$.
To improve coordination, a second-order module scores concept pairs $(c_{i,k},c_{j,m})$ ($j\neq i$) against $\boldsymbol{q}$, aggregates over other agents to form $\beta_{i,k}$, and normalizes across $k$.
Finally, ELV reweights concepts with $\alpha_{i,k}+\beta_{i,k}$ and takes a linear head to sum the individual value $Q_i(\tau_i^t,u_i^t)$.}
\vskip -0.1in
	\label{framework}

\end{figure}

\section{Method}
In this section, we detail the formulation of the \textit{Escaping Local Views} (ELV) framework within POMDP, which is designed to induce interpretable decision behaviors from opaque observation streams.
As illustrated in \textit{Figure}~\ref{framework}, to handle the partial observability, ELV seeks to reconcile the inherent conflict between local partial observability and the necessity for global coordination.
The framework is built upon three pillars: (1) establishing an interpretable semantic bottleneck that extracts disentangled concepts from local histories; (2) inducing a global latent bridge via a VAE to align disparate local views; and (3) incorporating a forward semantic dynamics model that utilizes prediction errors as intrinsic motivation, ensuring exploration is guided by prediction errors in the learned concept space rather than stochastic noise.

\subsection{Agent Concepts Network Design}
\label{sec:concept-predictor}
At each timestep $t$, each agent $i$ takes its local observation $o_i^t$ and the hidden state $h_i^{t-1}$, and generates a compact set of semantic concepts bottleneck $\mathcal{C}$ that outputs the local action values $Q_i$.
We maintain a recurrent history state $h_i^t=\mathrm{MLP}(h_i^{t-1},\, o_i^t,\, u_i^{t-1})$, where $u_i^{t-1}$ is the last action.
This recurrent mechanism allows the agent to retain information from experience, which is helpful for agents in non-stationary and partially observable environments.

The generated history state $h_i^t$ is then projected into $K$ concept slots, where each slot generates separate embeddings for the presence and absence of a semantic concept denoted by $\hat{c}_i^{+}(\boldsymbol{h})$ and $\hat{c}_i^{-}(\boldsymbol{h})$ using lightweight MLP heads $\phi_i^{+}$ and $\phi_i^{-}$.
To link the learned semantics with the target concept signal, we introduce a differentiable and learnable scoring function $\psi:\mathbb{C}^{2d}\rightarrow[0,1]$ that maps the joint dual-state embedding to a scalar existence probability.
Specifically, the probability $p_i$ is computed as:
\begin{equation}
p_i \triangleq \psi(\hat{c}_i^{+}(\boldsymbol{h}),\, \hat{c}_i^{-}(\boldsymbol{h}))^T = \sigma(\boldsymbol{w}_h(\hat{c}_i^{+}(\boldsymbol{h}),\, \hat{c}_i^{-}(\boldsymbol{h}))^T + \boldsymbol{b}_h),
\end{equation}
where $d$ denotes the concept dimension, $\sigma$ is the Sigmoid function, and $\boldsymbol{w}_h,\boldsymbol{b}_h$ are shared across all $K$ concepts to promote consistent scoring behavior and parameter efficiency.
Conditioned on $p_i$, the final concept embedding $c_i$ is obtained as a continuous mixture of the active and inactive embeddings:
\begin{equation}
c_i = p_i\, c_i^{+}(\boldsymbol{h}) + (1-p_i)\, c_i^{-}(\boldsymbol{h}).
\end{equation}
This formulation implements a soft gating mechanism in which $p_i$ functions as a differentiable switch that interpolates between active and inactive embeddings, yielding semantically distinct representations while preserving end-to-end differentiability for joint optimization.

To compensate for limited local views caused by partial observability, ELV constructs a global semantic bridge that distills the aggregated concept set $\boldsymbol{c}=[c_1;\dots;c_n]$ across all agents into a compact latent context vector $z$.
We approximate the posterior over this context with a variational distribution parameterized by neural networks $q_{\phi}(z\mid \boldsymbol{c})=\mathcal{N}\!\big(\mu(\boldsymbol{c}),\,\mathrm{diag}(\sigma^2(\boldsymbol{c}))\big)$, and maximize the ELBO by reconstructing the global state $s$ from $\boldsymbol{z}$ via $p_{\theta}(s\mid \boldsymbol{z})$ during centralized training:
\begin{equation}\label{F:vaeloss}
\mathcal{L}_{vae}
=\mathbb{E}_{q_{\phi}(\boldsymbol{z}\mid \boldsymbol{c})}\! [\log p_{\theta}(s\mid \boldsymbol{z})] -\beta\,D_{\text{KL}}\!(q_{\phi}(\boldsymbol{z}\mid \boldsymbol{c})\,\|\,p(\boldsymbol{z})),
\end{equation}
with a standard normal prior $p(\boldsymbol{z})=\mathcal{N}(0,I)$.
At execution, $\boldsymbol{z}$ is inferred solely from $\boldsymbol{c}$, providing the agent with global semantic context. Appendix~\ref{app:VAEs} provides an information-theoretic analysis.

\textbf{First- and Second-Order Concept Importance.}
Given the inferred global context $z$, we compute $1$-order concept importance via scaled dot-product attention between $z$ and each concept embedding:
\begin{equation}\label{F:importanceweight}
\alpha_k=\frac{\exp ((\boldsymbol{w}_z^1 \boldsymbol{z}+\boldsymbol{b}_z)^{\top}(\boldsymbol{w}_c^1 c_k+\boldsymbol{b}_1))}{\sum_j \exp ((\boldsymbol{w}_z^1 \boldsymbol{z}+\boldsymbol{b}_z)^{\top}(\boldsymbol{w}_c^1 c_j+\boldsymbol{b}_1))},
\end{equation}
where $\boldsymbol{w}_z^1$, $\boldsymbol{b}_z$, $\boldsymbol{w}_c^1$, $\boldsymbol{b}_1$ are learnable parameters.
The scalar $\alpha_k\in(0,1)$ quantifies how strongly the latent context activates concept $k$.
However, first-order attention alone may miss cross-concept dependencies critical for cooperation.
To address this, we introduce a second-order interaction mechanism that aggregates interaction affinities between a given concept and all external concepts $\tilde{\boldsymbol{c}}$:
\begin{equation}\label{eq:second_attn}
\beta_k =
\frac{
  \exp\!\left(
    \sum_{c_m\in\tilde{\boldsymbol{c}}}
      (\boldsymbol{w}_z^2 z+\boldsymbol{b}_z)^\top
      (\boldsymbol{w}_c^2[c_k\Vert c_m] + \boldsymbol{b}_2)
  \right)
}{
  \sum_j
  \exp\!\left(
    \sum_{c_m\in\tilde{\boldsymbol{c}}}
      (\boldsymbol{w}_z^2 z+\boldsymbol{b}_z)^\top
      (\boldsymbol{w}_c^2[c_j\Vert c_m] + \boldsymbol{b}_2)
  \right)
},
\end{equation}
where $[\,\cdot \| \cdot\,]$ denotes concatenation and $\boldsymbol{w}_2$, $\boldsymbol{b}_2$ are learnable parameters.
The 2-order weight $\beta_k$ highlights concepts that are not only globally relevant but also beneficial for coordination.
1-order and 2-order signals are fused to reweight concept embeddings, and the action value is computed as:
\begin{equation}\label{Eq:computeQ}
Q_i(\tau_i, u_i)= \boldsymbol{w}_Q(\sum\nolimits_k(\alpha_i^k+\beta_i^k)\,c_i^k) + \boldsymbol{b}_Q,
\end{equation}
where $\boldsymbol{w}_Q$ and $\boldsymbol{b}$ are learnable parameters that map the aggregated concepts to Q-values.
In this formulation, the combined importance scores $\alpha_{i,k}+\beta_{i,k}$ allow the model to attend dynamically to both contextual relevance and interaction relevance, which can help improve coordination and value estimation in cooperative MARL.
The theoretical analysis for the concept importance with $\alpha$ and interaction-aware importance with $\beta$ is provided in Appendix~\ref{sec:appendix_theory}.

\subsection{Reconstruction Loss}
While the concept bottleneck supports compact, interpretable representations, unconstrained concept embeddings may drift into arbitrary latent codes which lack meaningful observational grounding. 
To counteract this, we introduce an auxiliary reconstruction objective that explicitly anchors each agent’s concept embedding to its local observation space. 
A lightweight decoder $\mathcal{D}$ predicts the original observation $o_i^t$ from the corresponding concept embedding $c_i^t$, effectively constraining the learned concepts to retain task-relevant perceptual information and preventing representational collapse:
\begin{equation}\label{F:recloss}
\mathcal{L}_{rec} = \mathbb{E}_t [ |\mathcal{D}(c_i^t) - o_i^t |_2^2 ].
\end{equation}

\subsection{Intrinsic Reward by Predicting Concepts}
Efficient exploration remains a central challenge in reinforcement learning, particularly in complex multi-agent environments where rewards are sparse or delayed.
To address this, we propose a concept-driven exploration mechanism that generates intrinsic rewards based on semantic dynamics rather than raw perceptual signals.
Unlike intrinsic rewards defined directly on observations, which are prone to irrelevant perceptual noise, our method evaluates novelty in a structured concept space to drive exploration toward semantically meaningful transitions.

We instantiate a forward dynamics model $f_\theta$ over concept embeddings.
At each timestep $t$, given agent $i$'s concept embedding $c_i^t$ and executed action $a_i^t$, we predict the next concept embedding:
\begin{equation}
\hat{c}_i^{t+1} = f_\theta(c_i^t, a_i^t).
\end{equation}
And the forward model is trained to minimize the forward prediction loss
\begin{equation} 
\mathcal{L}_{pred} = \mathbb{E}_t[|\hat{c}_i^{t+1}-c_i^{t+1}|_2^2],
\end{equation}
which helps the predictor models temporal dependencies of concept evolution.
The overall grounding objective $\mathcal{L}_{G_w}$ jointly minimizes the variational semantic loss $\mathcal{L}_{vae}$ and the reconstruction loss $\mathcal{L}_{rec}$:
\begin{equation}
 \mathcal{L}_{G_w}=\lambda_1\mathcal{L}_{vae}+ \lambda_2\mathcal{L}_{rec}+\lambda_3\mathcal{L}_{pred},
\end{equation}
where $\lambda_1$ encourages the latent context $z$ to preserve global semantics via state reconstruction, $\lambda_2$ ensures that local concepts remain predictive of perceptual inputs, and $\lambda_3$ governs the learning of the forward dynamics model, ensuring the concept representation captures the temporal evolution and causal structure of the environment.
In information-theoretic terms, minimizing $\mathcal{L}_{\text{rec}}$ serves to maximize a lower bound on the mutual information between the concept embedding $c_i^t$ and the local observation $o_i^t$. 
Mutual information quantifies the shared information between two variables, i.e., how much knowing one reduces uncertainty about the other, and is widely used in representation learning to encourage representations to retain informative, task-relevant structure rather than noise.
Maximizing this dependency drives concept embeddings to more faithfully encode semantically meaningful features of the observation.
A formal derivation showing that reducing the reconstruction loss increases this mutual information lower bound is provided in Appendix~\ref{app:decoder}.

\textbf{Semantic Novelty as Intrinsic Reward}. 
We interpret the prediction error as a measure of semantic novelty: large prediction residuals indicate transitions that are unfamiliar or poorly understood.
Formally, we define the intrinsic reward as the next-concept prediction error across all agents:
\begin{equation}\label{F:intrinsic reward}
r_{t}^{\text{int}} = |\hat{\boldsymbol{c}}^{t+1} - \boldsymbol{c}^{t+1}|_2^2.
\end{equation}
This formulation aligns with prediction-error-based intrinsic motivation, where the prediction error of a learned forward predictor functions as a novelty signal that encourages exploration of under-modeled dynamics.
Such curiosity-driven rewards have been widely shown to improve exploration efficiency by pushing agents to sample transitions with high prediction error, especially under sparse extrinsic feedback.
By deriving intrinsic rewards in the learned concept space rather than the raw observation space, the exploration signal encourages exploration through concept-space prediction errors and remains focused on semantically meaningful transitions that are aligned with task structure.

To balance exploration and exploitation, we anneal the intrinsic component over the course of training $r_{t} = r_{t}^{\text{env}} + \beta_t \cdot r_{t}^{\text{int}}$, with $\beta_t = \beta_0 \cdot \exp(-\lambda t)$, where $\beta_0$ and $\lambda$ control the initial scale and decay rate of the intrinsic signal, respectively. 
This schedule biases early learning toward structured semantic exploration when the model's uncertainty is high and progressively shifts emphasis to maximizing extrinsic task rewards as the dynamics become better learned.

\subsection{Overall Learning Objective}
Under CTDE, we derive each agent's individual value from its concept representations and aggregate them using a value factorisation network (e.g., QMIX~\citep{rashid2020monotonic} or VDN~\citep{sunehag2018value}) to form the joint action-value function $Q_{\mathrm{tot}}$.
We optimise the value estimator by minimizing squared temporal-difference (TD) error over samples from a replay buffer $\mathcal{B}$:
\begin{gather}\label{eq:td_loss}
\mathcal{L}_{TD}(\theta)=\mathbb{E}_{(\tau,u,r,\tau')\sim\mathcal{B}}[(y-Q_{\mathrm{tot}}(\boldsymbol{\tau},\boldsymbol{u};\theta))^2],
\end{gather}
where $y=r+\gamma\max_{\boldsymbol{u}'}Q_{\mathrm{tot}}(\boldsymbol{\tau'},\boldsymbol{u'};\theta^-)$ is the one-step Bellman target computed with the target network parameters $\theta^-$ copied from $\theta$ periodically.
The overall loss jointly optimises value estimation and semantic representation learning as \begin{equation}\label{eq:total_loss}
\mathcal{L}(\theta,w)
=\mathcal{L}_{TD}
+\mathcal{L}_{G_w}.
\end{equation}
This joint optimisation aligns accurate value estimation with semantically informed concept learning, ensuring that the learned representations meaningfully support coordinated decision making.
We summarize the pseudo-code of our training procedure in Appendix~\ref{pseudocode}.

\section{Experiments}\label{Experiments}
We evaluate ELV on three challenging benchmarks: SMAC, SMACv2, and Level-Based Foraging (LBF).
Baselines include (i) classic value decomposition methods (VDN, QMIX, QPLEX~\citep{wang2021qplex}, QTRAN~\citep{son2019qtran}, CDS~\citep{li2021celebrating}) and (ii) recent CTDE approaches that strengthen per-agent value estimation and credit assignment via architectural or objective-level enhancements (NA$^2$Q~\citep{liu2023na2q}, ReBorn~\citep{qin2024dormant}, MA$^2$E~\citep{kangma}, SHAQ~\citep{wang2022shaq}).
Implementation details and benchmark configurations are provided in Appendix~\ref{ExperimentalDetails}.
All performance curves report the mean $\pm$ standard deviation over five random seeds.

\subsection{Performance on LBF}
\begingroup
\setlength{\intextsep}{6pt}
\begin{wrapfigure}{r}{0.55\textwidth}
  \centering
  \paperfigure{\linewidth}{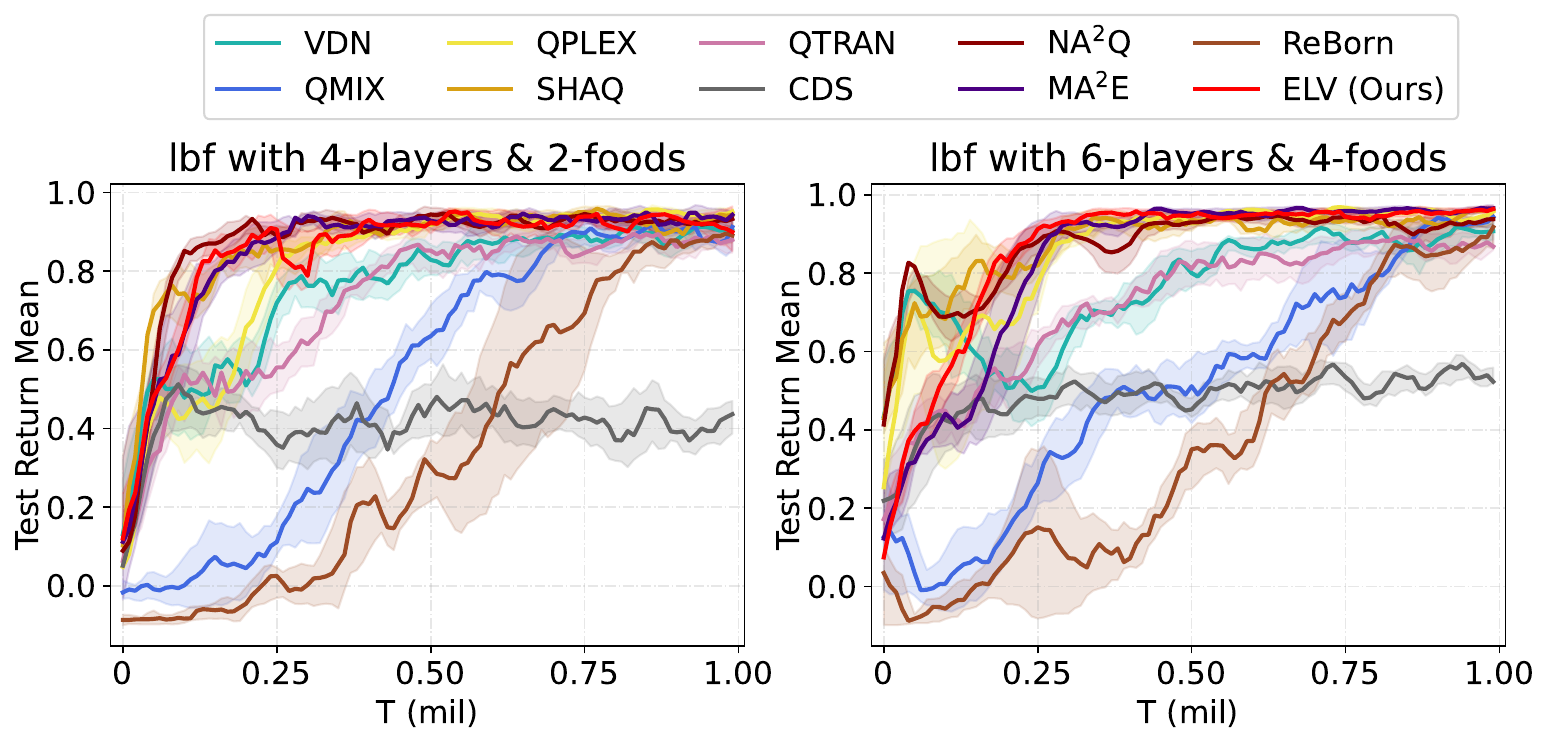}
  \vskip -0.1in
  \caption{Performance comparison on two tasks of LBF.}
  \label{lbf}
\end{wrapfigure}
We first evaluate ELV on two constructed LBF settings.
We instantiate ELV on top of the QMIX value-factorisation backbone—keeping the QMIX mixing network unchanged and replacing only the individual value module with ELV.
As shown in \textit{Figure}~\ref{lbf}, ELV achieves better test returns with competitive sample efficiency and stable learning dynamics.
VDN exhibits delayed progress, which may be consistent with its limited capacity to represent complex joint action-values under coupled credit assignment.
CDS plateaus at a substantially lower return, suggesting that encouraging diversity at the individual level may come at the expense of coherent team strategies in tightly coupled coordination.
ReBorn also converges late, indicating inefficient exploration in the expanded joint strategy space. 
Compared with QMIX, ELV yields a clear performance gain while keeping the same mixing network fixed. 
This indicates that the improvement primarily comes from the individual-value architecture: replacing the GRU-based utility with a concept-based representation of ELV better mitigates partial observability and leads to more consistent coordination.
\par\endgroup

\begin{figure}[tb]
\centering

		\paperfigure{0.95\textwidth}{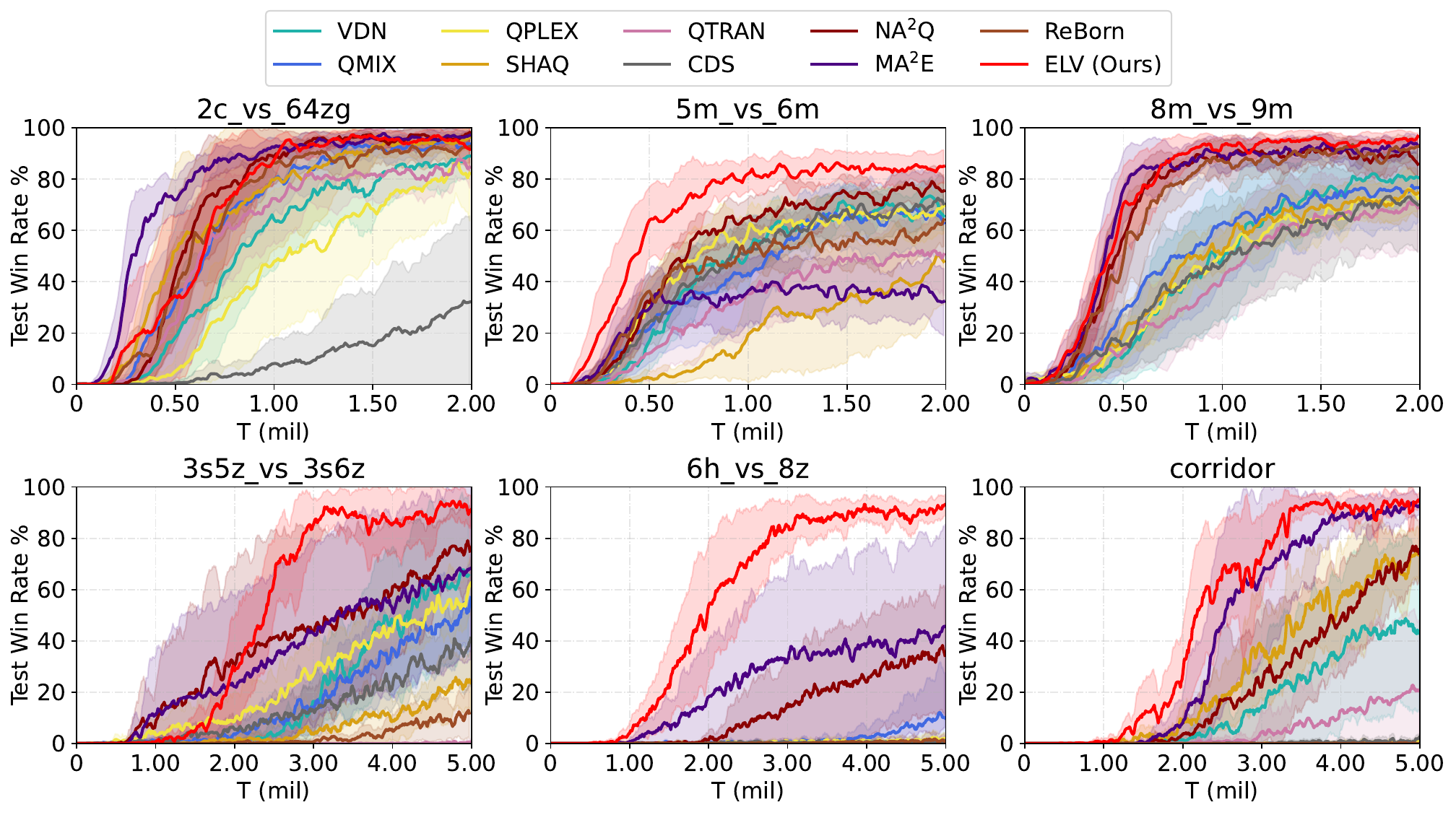}
	\vskip -0.1in
	\caption{Performance comparison on hard and super-hard scenarios.}
	\vskip -0.1in
	\label{3hard3superhard}

\end{figure}

\begin{figure}[htbp]
\centering
		\paperfigure{0.95\textwidth}{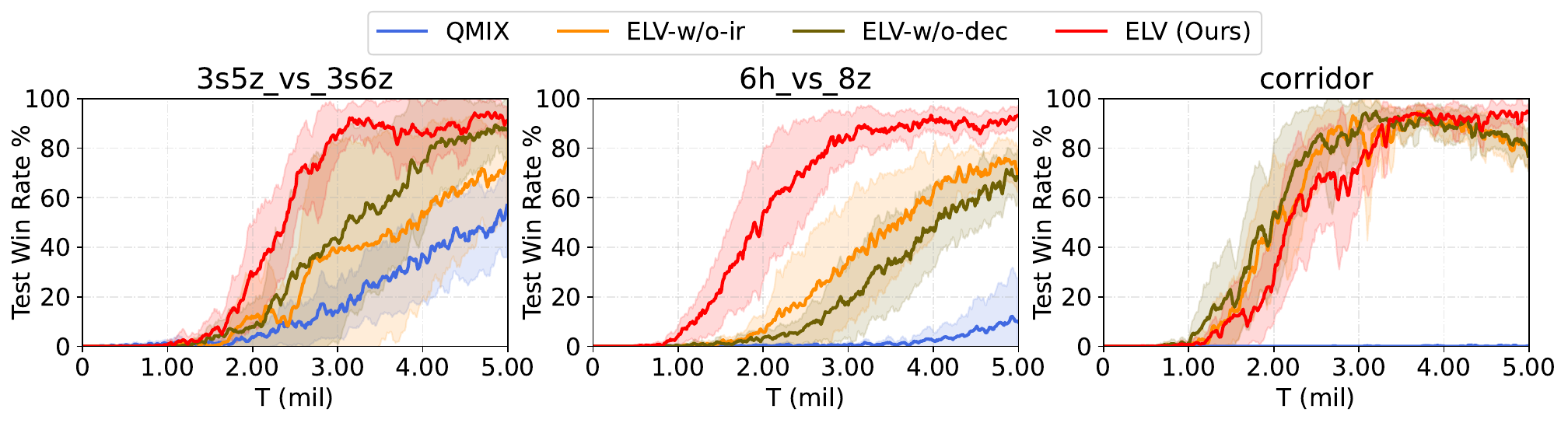}
	\vskip -0.1in
	\caption{Ablation of intrinsic reward and concept decoder for ELV.}
	\vskip -0.2in
	\label{ablation}
\end{figure}

\subsection{Performance on SMAC}
For the SMAC benchmark, we evaluate ELV on a diverse set of maps spanning easy, hard, and super-hard scenarios.
As shown in \textit{Figure}~\ref{3hard3superhard}, ELV achieves consistently stronger win rates, with the clearest margins on super-hard tasks where partial observability and long-horizon coordination are most limiting.
For example, on \textit{3s5z\_vs\_3s6z} and \textit{6h\_vs\_8z}, ELV continues improving after QMIX or VDN largely plateau and stabilises at a high win rate, matching the visual trend that ELV both converges faster and sustains better asymptotic performance in tightly coupled fights.
These patterns are consistent with known limitations of value factorisation baselines: VDN's additive form and QMIX's monotonic mixing can limit the expressivity of $Q_{\mathrm{tot}}$ in highly coupled settings, often leading to early saturation on harder maps.
QTRAN relaxes these constraints but depends on stricter optimisation objectives that can be brittle at scale, which aligns with its weaker performance in the challenging regime.
CDS promotes behavioural diversity, but on maps requiring precise team-level coordination, this does not necessarily translate into coherent joint strategies and can plateau below top methods.
While MA$^2$E outperforms most baselines, it still trails ELV, with the gap most pronounced on the super-hard tasks.
It suggests that each agent has the ability to infer the information of other agents from partial observations.
Compared to VDN and QMIX, ELV yields larger gains, which we attribute to its semantically structured individual-value estimator: the concept bottleneck together with the variational semantic bridge mitigates information fragmentation under partial observability, enabling more consistent coordination than a GRU-based individual value function while keeping the same mixing network.
More results of plugging ELV into VDN and QMIX backbones are reported in Appendix~\ref{elvvdn} and ~\ref{extrasmacqmix}.
Additional experiments on SMACv2 are provided in Appendix~\ref{PerformanceonSMACv2}.

\subsection{Ablation Study}
\label{sec:ablation}

\begin{figure}[tb]
\centering
		\paperfigure{0.92\textwidth}{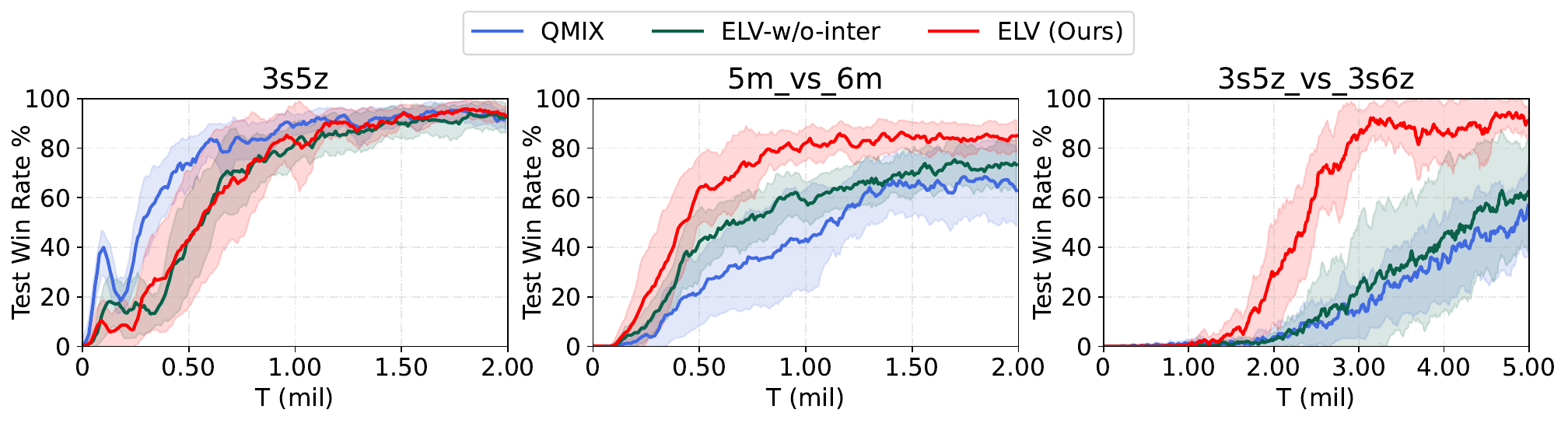}
	\vskip -0.1in
	\caption{Ablation of the concept interaction module for ELV.}
	\vskip -0.1in

	\label{ELV_wo_inter}
\end{figure}

\begin{figure}[tb]
\centering
		\paperfigure{0.92\textwidth}{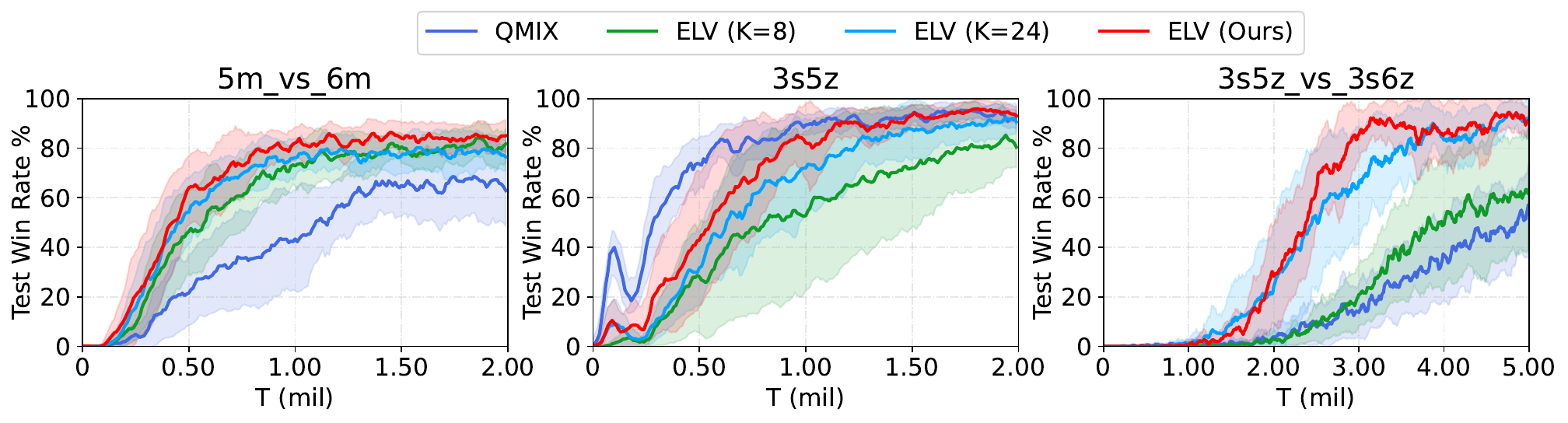}
	\vskip -0.1in
	\caption{Ablation study with different numbers of concepts for ELV.}
	\vskip -0.2in
	\label{ELV_diff_concept}
\end{figure}
We perform ablations on the three core components of ELV—concept-dynamics intrinsic reward (ir), reconstruction-based grounding (dec), and concept-level interactions (inter)—by evaluating the corresponding variants without each component (ELV-w/o-ir, ELV-w/o-dec, and ELV-w/o-inter).

As shown in \textit{Figure}~\ref{ablation}, when removing the concept-dynamics intrinsic reward causes a sharp drop on \textit{3s5z\_vs\_3s6z}, ELV-w/o-ir gets stuck in poor local solutions, whereas the full model exceeds $90\%$ win rate.
A similar effect appears on \textit{Corridor}, where dropping ir slows learning and lowers final performance, supporting the role of concept-space prediction error as an effective exploration signal under sparse or misleading rewards.
In contrast, the reconstruction grounding mainly matters under heterogeneity on \textit{6h\_vs\_8z}.
ELV without dec degrades the most, which suggests that grounding stabilizes concept semantics across diverse unit types, while on the navigation-heavy \textit{Corridor}, its impact is minor.
To isolate the effect of interaction from task difficulty, we compare the interaction-removed variant with the full model as shown in \textit{Figure}~\ref{ELV_diff_concept}.
The benefit is that the interaction module yields only minor gains on the simpler \textit{3s5z} map, but the gap grows markedly on the super-hard \textit{3s5z\_vs\_3s6z}.
This pattern suggests that explicit concept-level coordination becomes more important as the joint strategy space expands and tighter cooperation is required.

We finally study how concept capacity interacts with scalability by varying the number of concepts $k \in {8,16,24}$ across scenarios.
As shown in \textit{Figure}~\ref{ELV_diff_concept}, we can find that higher concept capacity is required to maintain both final performance and sample efficiency as the number of agents grows.
The compact setting ($k=8$) forms a clear bottleneck on complex maps such as \textit{3s5z\_vs\_3s6z}, where performance drops sharply, whereas the larger setting ($k=24$) shows better scalability: it brings little benefit on small maps but yields markedly more robust returns in high-complexity environments.
These results indicate that a sufficient number of concept slots is crucial for representing the rich joint strategies needed in large-scale multi-agent coordination.

\subsection{Interpretability}
\label{Explanation}
\begingroup
\setlength{\intextsep}{6pt}
\begin{wrapfigure}[15]{R}{0.42\textwidth}
  \centering
  \paperfigure{\linewidth}{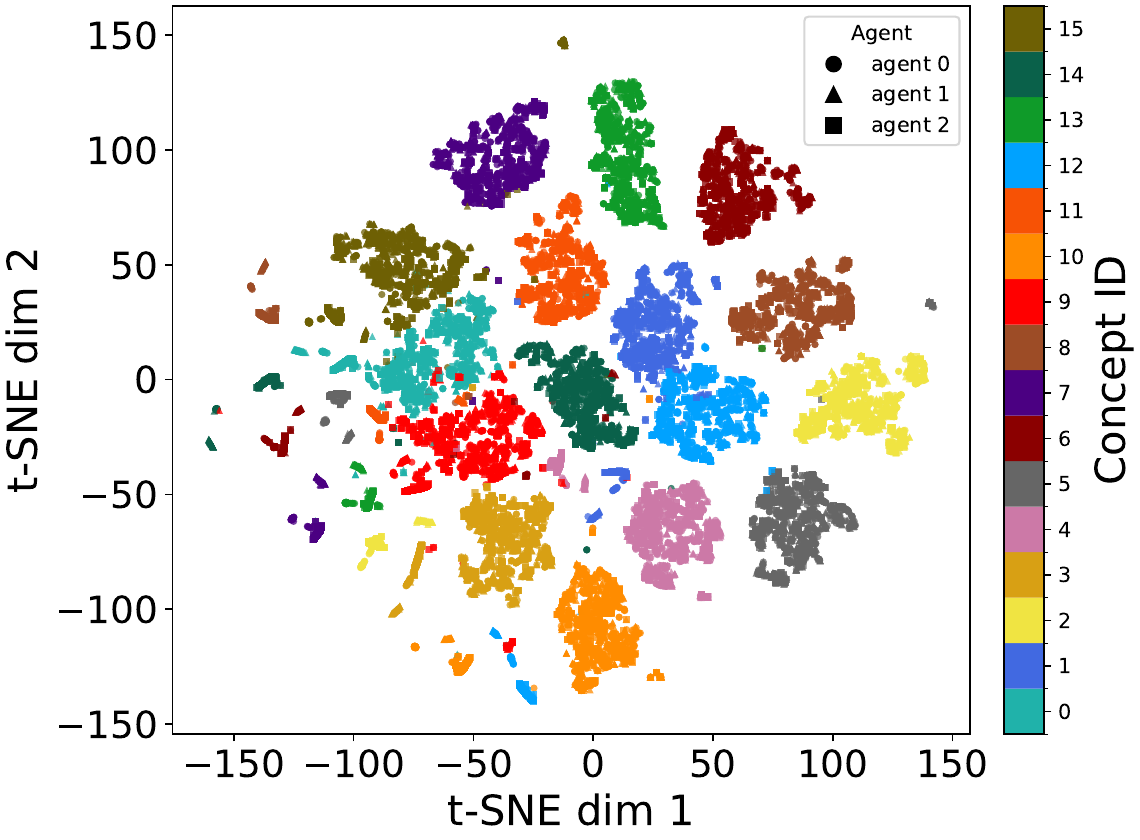}
  \caption{The t-SNE visualization of learned concept embeddings.}
  \label{tsne}
\end{wrapfigure}
\textbf{Concept Space Visualization.} To examine the structure of the learned concepts, we visualize their embeddings with t-SNE~\citep{maaten2008visualizing} in \textit{Figure}~\ref{tsne}.
The embeddings form well-separated, compact clusters (colors), indicating that ELV disentangles the state space into distinct semantic factors.
Moreover, points from different agents (shapes) are well-mixed within each cluster, which suggests that the concepts are agent-invariant and can be shared across the team.

\textbf{Grounding Concepts via Gradient-based Attribution.}
To further clarify concept semantics, we adopt a gradient-based attribution scheme at evaluation time to relate each concept to structured observation modalities (e.g., movement directions, ally/enemy states, self attributes).
Concretely, we compute the attribution of concept $c_i$ to observation modality $f_j$, which yields a concept–feature attribution matrix $G \in \mathbb{R}^{C \times F}$ with entries $G_{i j}:=\|\frac{\partial \operatorname{Recon}_{f_j}}{\partial c_i}\|_1$.
It measures how sensitive the reconstruction of feature $f_j$ is to perturbations in concept $c_i$ and thus serves as a proxy for their semantic influence.

To obtain a concise and human-readable assignment, we then solve a maximum-attribution matching problem that assigns each observation modality to a single concept $\max _\pi \sum_{j=1}^F G_{\pi(j), j}$, $\text { s.t. } \pi:\{1, \ldots, F\} \rightarrow\{1, \ldots, C\} \text { injective }$, where $G_{ij}$ denotes the attribution strength from concept $i$ to feature $j$, and the injective mapping $\pi$ ensures that each feature is assigned to exactly one concept.
\textit{Figure}~\ref{concept assignment} (right) illustrates the resulting heatmap.
The clear high-intensity blocks indicate that ELV disentangles the raw observation space, mapping different latent concepts to distinct semantic groups (e.g., specific allies or enemies) rather than learning an entangled representation.

\begin{figure}[!t]
\centering
	\paperfigure{1.0\textwidth}{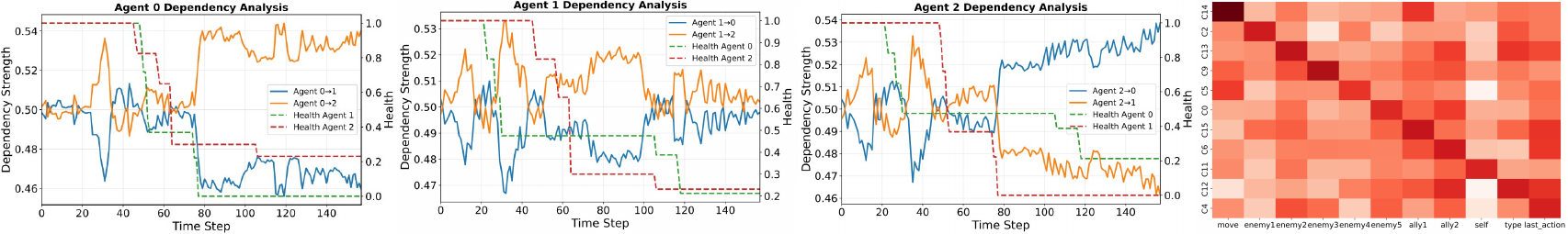}
	\vskip -0.1in
	\caption{Dependency strength over health and concept assignment heatmap.}
	\vskip -0.1in
	\label{concept assignment}
\end{figure}
\begin{figure}[!t]
\centering
	\paperfigure{0.9\textwidth}{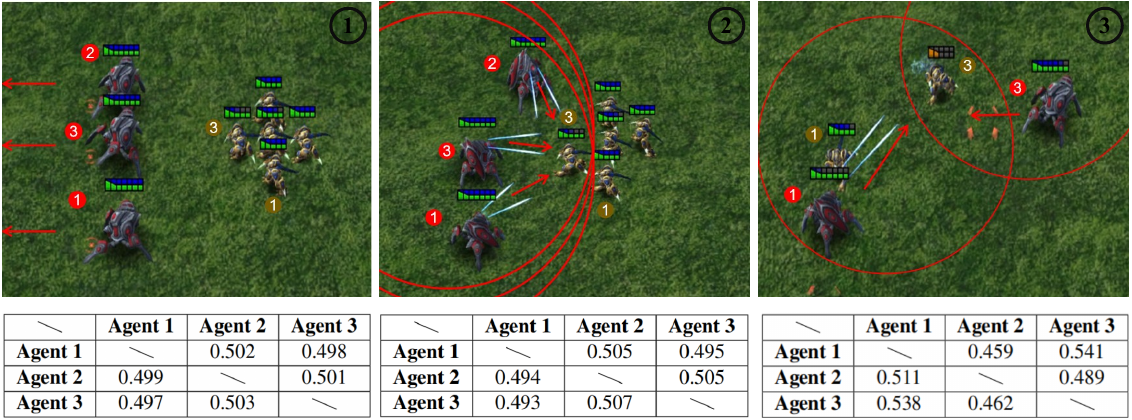}
	\vskip -0.1in
	\caption{Visualization of agent collaboration dynamics.}
	\vskip -0.2in
	\label{explain1}
\end{figure}

\textbf{Qualitative Case Study.} To quantify how ELV models cooperative coupling between agents, we define an \emph{agent dependency strength} $\mathcal{D}_{ij}$ based on the learned concept–interaction scores.
For each pair of agents $i \neq j$, we first aggregate interaction weights over all pairs of latent concepts $\mathcal{M}_{ij} = \sum \sum m(c_i^{k_1}, c_j^{k_2})$, $m(c_i^{k_1}, c_j^{k_2}) = (\boldsymbol{w}_z \boldsymbol{z}+\boldsymbol{b}_z)^{\top} (\boldsymbol{w}_2 [c_i^{k_1} \,\|\, c_j^{k_2}] + \boldsymbol{b}_2)$, and then normalize via a row-wise softmax to get a relative dependency distribution $\mathcal{D}_{ij}=\frac{\exp(\mathcal{M}_{ij})}{\sum_j \exp(\mathcal{M}_{ij})}$.
Here, $\mathcal{D}_{ij}$ measures how agent $i$ depends on agent $j$ as it aggregates concept-level interactions.

\textit{Figure}~\ref{explain1} illustrates three representative phases from a SMAC rollout.
At the beginning of the episode, concept activations are relatively uniform, and no specific tactical target stands out.
Accordingly, dependency scores between agents remain close to a balanced split (e.g., Agent~1 allocates its attention to teammates roughly $0.502/0.498$), suggesting that only a relatively balanced allocation of attention across teammates is required for simple coordinated retreat.
In the middle phase, an abrupt increase in the activation of a particular concept (Concept~9, associated in this run with a specific enemy unit) coincides with the onset of a coordinated counter-attack.
Although the agents move into a tightly clustered formation, the dependency matrix stays relatively stable, indicating that once a common tactical focus emerges, ELV maintains coherent group behavior without needing large shifts in pairwise attention.
In the final phase, agents split spatially: Agent~3 chases `Enemy~3' while Agent~1 provides support from a distance.
Here, we observe a clear dependency asymmetry: the dependency of Agent~1 on Agent~3 rises to $0.541$, while its dependency on nearby agents decreases.
This pattern shows that ELV prioritizes concept-level interaction (who engages the key target) over mere spatial proximity, enabling consistent joint tactics under physical dispersion.
Tracking $\mathcal{D}_{ij}$ over time further reveals that dependencies on a given teammate tend to decrease as that agent's health drops, consistent with \textit{Figure}~\ref{concept assignment}, suggesting that ELV gradually down-weights coordination with vulnerable agents to stabilize the overall policy.

\par\endgroup

\section{Conclusion}
We presented an interpretable cooperative MARL framework with a shared concept bottleneck. ELV decomposes local observations and histories into semantically structured latent concepts and aligns them via a variational semantic bridge, thereby mitigating partial observability and improving global coordination. Dual-path attention captures context-dependent salience and higher-order concept interactions, while reconstruction-based grounding and concept-dynamics intrinsic rewards regularise representations and promote exploration. Across a range of MARL benchmarks, ELV consistently attains competitive or superior performance relative to strong baselines and yields concept-level explanations of coordination strategies.
By coupling value estimation with explicit semantic structure, ELV advances transparent, data-efficient multi-agent systems in complex domains.

\subsection*{AI use statement}
In preparing this manuscript, we used generative AI tools to improve the readability and concision of the text and to assist with LaTeX formatting checks. All AI-assisted revisions were reviewed by the authors, who take full responsibility for the final content of this work.

\bibliography{example_paper}
\bibliographystyle{iclr2027_conference}

\clearpage
\appendix

\section{Theoretical Justification for Hierarchical Concept Modeling}
\label{sec:appendix_theory}

In this section, we provide theoretical grounding for explicitly modeling first-order concept importance (individual semantics) and second-order concept interactions (pairwise synergies) within the ELV framework. Our justification is based on the functional ANOVA (Hoeffding-style) decomposition together with a second-order truncation as an inductive bias. This perspective clarifies that our architecture is not merely heuristic: it corresponds to a structured approximation of a general value (or policy) function defined over a learned concept space.

\subsection{Functional ANOVA--Style Decomposition}
\label{subsec:anova}

Consider the learned action-value function as a multivariate function defined on the joint concept variables. Let $\mathbf{C}=(C_1,\ldots,C_M)$ denote the set of all concepts extracted across agents (under the data distribution induced by the environment and the current policy), and let $f(\mathbf{C},a)$ represent the target function (e.g., a per-agent $Q$-function or a utility component) conditioned on action $a$.

A classical result in functional ANOVA (often attributed to Hoeffding) states that, for $f(\cdot,a)\in L_2(\mathbb{P}_{\mathbf{C}})$, one can express $f$ as a sum of components of increasing order,
\begin{equation}
f(\mathbf{C},a)
= f_0(a)
+ \sum_{i=1}^{M} f_i(C_i,a)
+ \sum_{1\le i<j\le M} f_{ij}(C_i,C_j,a)
+ \cdots,
\label{eq:anova_expansion}
\end{equation}
where $f_0(a)$ is a global baseline and $f_i$, $f_{ij}$, etc., capture effects attributable to individual concepts and their interactions, respectively. To make these components \emph{identifiable}, a common choice is to impose centering (purification) constraints, e.g.,
\begin{equation}
\mathbb{E}\!\left[f_i(C_i,a)\right]=0,\qquad 
\mathbb{E}\!\left[f_{ij}(C_i,C_j,a)\mid C_i\right]=0,\qquad
\mathbb{E}\!\left[f_{ij}(C_i,C_j,a)\mid C_j\right]=0,
\label{eq:anova_centering}
\end{equation}
with expectations taken under $\mathbb{P}_{\mathbf{C}}$. Under additional conditions (e.g., a product measure over inputs), these components are orthogonal in $L_2$; in general, we adopt this decomposition as a modeling lens and structural prior that separates main semantic effects from pairwise semantic synergies.

Within this view, $f_i(C_i,a)$ corresponds to the first-order (main) effect of concept $C_i$ on the decision, aligning with our first-order attention branch. The term $f_{ij}(C_i,C_j,a)$ corresponds to the second-order interaction effect, capturing the residual dependence on $(C_i,C_j)$ beyond what is explained by the individual main effects, aligning with our interaction branch.

\subsection{Second-Order Truncation as an Inductive Bias}
\label{subsec:truncation}

While the full expansion includes higher-order interactions up to order $M$, explicitly modeling all terms is statistically and computationally prohibitive: the number of possible components grows exponentially with $M$ (on the order of $2^{M}$), and higher-order components are typically harder to estimate reliably from finite data.

ELV therefore truncates the expansion at second order and approximates $f$ by
\begin{equation}
f(\mathbf{C},a)\approx f_0(a)+\sum_{i=1}^{M} f_i(C_i,a)+\sum_{1\le i<j\le M} f_{ij}(C_i,C_j,a).
\label{eq:second_order_trunc}
\end{equation}
This choice should be interpreted as a deliberate inductive bias: it enforces a transparent hierarchy in which (i) individual concepts provide interpretable semantic evidence for an action, and (ii) pairwise interactions capture salient synergies that cannot be expressed by purely additive effects. Importantly, we do \textbf{not} claim that higher-order interactions are absent in principle; rather, the second-order truncation provides a favorable trade-off between expressivity, robustness, and interpretability, which we validate empirically. This is particularly appropriate for coordination patterns that are naturally pairwise (e.g., complementary roles or two-factor conditions that jointly trigger a behavior).

\subsection{Connection to Coordination Graphs}
\label{subsec:coord_graphs}

Our decomposition is also conceptually aligned with coordination-graph-based factorization in cooperative MARL, where a joint utility function is expressed as a sum of unary and pairwise potentials over graph nodes and edges. ELV instantiates an analogous unary + pairwise structure, but crucially applies it in the latent semantic (concept) space rather than directly over agent identities. As a result, ELV can capture semantic interactions (e.g., ``low health'' interacting with ``enemy proximity'') regardless of which agent expresses these concepts, supporting improved interpretability while retaining sufficient capacity for complex decision-making.

\section{Theoretical Justification for Concept-based Reconstruction}
\label{app:decoder}

In this section, we provide the theoretical justification for the auxiliary reconstruction objective. We demonstrate that minimizing the reconstruction loss is equivalent to maximizing a variational lower bound of the Mutual Information (MI) between the local observation and the extracted concept, under the empirical distribution induced by the current policy.

\subsection{Problem Formulation}

Let $O$ be the continuous random variable representing the local observation, and $C$ be the continuous random variable for the concept representation. The concept extractor defines a (possibly stochastic) conditional distribution $p_\phi(c|o)$ (e.g., by injecting noise or outputting a parametric distribution), and we aim to maximize the Mutual Information $I(O; C)$:

\begin{equation}
    I(O; C) = H(O) - H(O|C),
\end{equation}

where $H(\cdot)$ denotes the differential entropy.

\subsection{Addressing Non-Stationarity in RL}

In Reinforcement Learning, the distribution of observations $p_\pi(o)$ is non-stationary as it depends on the evolving policy $\pi$. However, at any given training step $t$, we optimize the representation with respect to the empirical distribution $p_{\text{emp}}(o)$ (e.g., sampled from a replay buffer or a fixed minibatch). Conditioned on this empirical distribution, the marginal entropy term $H(O)$ can be treated as a constant with respect to the representation parameters within each optimization step. Our goal thus simplifies to minimizing the conditional differential entropy $H(O|C)$.

\subsection{Derivation of the Lower Bound}

The conditional differential entropy is defined as:
\begin{equation}
    H(O|C) = - \mathbb{E}_{p_{\text{emp}}(o,c)} [\log p_{\text{emp}}(o|c)],
\end{equation}
where $p_{\text{emp}}(o,c) = p_{\text{emp}}(o) p_\phi(c|o)$.

Here, $p_{\text{emp}}(o|c)$ denotes the conditional distribution induced by the empirical joint distribution, which is generally not available in closed form. To approximate it, we introduce a variational decoder $q_\theta(o|c)$.

Using the non-negativity of the Kullback-Leibler (KL) divergence, we have:
\begin{equation}
    D_{\text{KL}}(p_{\text{emp}}(\cdot|c) \,\|\, q_\theta(\cdot|c)) \ge 0,
\end{equation}
where
\begin{equation}
\begin{aligned}
    D_{\text{KL}}(p_{\text{emp}}(\cdot|c) \,\|\, q_\theta(\cdot|c))
    &= \mathbb{E}_{p_{\text{emp}}(o|c)}
    \left[\log \frac{p_{\text{emp}}(o|c)}{q_\theta(o|c)}\right] \\
    &= \mathbb{E}_{p_{\text{emp}}(o|c)}[\log p_{\text{emp}}(o|c)]
     - \mathbb{E}_{p_{\text{emp}}(o|c)}[\log q_\theta(o|c)].
\end{aligned}
\end{equation}
Therefore, $D_{\text{KL}}(\cdot) \ge 0$ implies 
$\mathbb{E}_{p_{\text{emp}}(o|c)}[\log p_{\text{emp}}(o|c)] \ge \mathbb{E}_{p_{\text{emp}}(o|c)}[\log q_\theta(o|c)]$. 
Taking the expectation over concepts $c$ extends this inequality to the joint distribution $p_{\text{emp}}(o, c)$:

\begin{equation}
    \mathbb{E}_{p_{\text{emp}}(o,c)} [\log p_{\text{emp}}(o|c)] \ge \mathbb{E}_{p_{\text{emp}}(o,c)} [\log q_\theta(o|c)].
\end{equation}

Substituting this inequality into the MI equation yields:
\begin{equation}
\begin{aligned}
    I(O; C) &= H(O) + \mathbb{E}_{p_{\text{emp}}(o,c)} [\log p_{\text{emp}}(o|c)] \\
    &\ge H(O) + \mathbb{E}_{p_{\text{emp}}(o,c)} [\log q_\theta(o|c)].
\end{aligned}
\end{equation}

We define the variational lower bound objective $\mathcal{L}_{\text{MI}}$ as:
\begin{equation}
    \mathcal{L}_{\text{MI}} \triangleq H(O) + \mathbb{E}_{p_{\text{emp}}(o,c)} [\log q_\theta(o|c)].
\end{equation}
Maximizing $\mathcal{L}_{\text{MI}}$ maximizes a lower bound on $I(O; C)$.

\subsection{Equivalence to Reconstruction Loss}

Since our decoder is a deterministic neural network $\hat{o} = g_\theta(c)$, we interpret it probabilistically by assuming $q_\theta(o|c)$ follows a fixed-variance Gaussian distribution centered at $\hat{o}$:
\begin{equation}
    q_\theta(o|c) = \mathcal{N}(o; g_\theta(c), \sigma^2 \mathbf{I}),
\end{equation}
where $\sigma^2$ is a fixed constant and $\mathbf{I}$ is the identity matrix. The log-likelihood term becomes:
\begin{equation}
    \log q_\theta(o|c) = -\frac{1}{2\sigma^2} \| o - g_\theta(c) \|^2 + \text{const}.
\end{equation}

Consequently, maximizing the MI lower bound is equivalent (up to an additive constant and a positive scaling factor) to minimizing the Mean Squared Error (MSE) reconstruction loss:
\begin{equation}
    \mathcal{L}_{\text{rec}} = \mathbb{E}_{p_{\text{emp}}(o,c)} \left[ \| o - g_\theta(c) \|^2 \right].
\end{equation}

\textbf{Conclusion} The reconstruction loss serves as a proxy for maximizing a variational lower bound on the mutual information between observations and concepts under the current empirical data distribution. This promotes faithful concept extraction by encouraging the concepts to retain information sufficient for reconstructing the local observations (while additional bottlenecks or regularizers, when used, can further discourage trivial copying and encourage compact representations).

\section{Enforcing Concept Sufficiency via Variational Information Bottleneck}
\label{app:VAEs}

In this appendix, we justify our auxiliary global-state reconstruction objective.
We consider the architecture where extracted concepts $c$ are compressed into a latent bottleneck $z$,
which is then used to reconstruct the global state $s$.
We show that minimizing the reconstruction loss maximizes a variational lower bound of $I(s;z)$, and by the
Data Processing Inequality (DPI) this pushes up a certified lower bound of $I(s;c)$.

\subsection{Graphical Model and Assumptions}

Let $s\in\mathcal S$ denote the global state and $c\in\mathcal C$ denote the aggregated concept representation
(e.g., $c=\{c_i\}_{i=1}^N$). We introduce a bottleneck latent $z\in\mathcal Z$ sampled from an encoder
$q_\phi(z\mid c)$, and reconstruct $s$ using a decoder $p_\theta(s\mid z)$.

\textbf{Assumption 1 (Markov property).}
Since $z$ is generated solely from $c$, the induced variables satisfy the Markov chain
\begin{equation}
  S \rightarrow C \rightarrow Z,
\end{equation}
equivalently $q_\phi(z\mid c,s)=q_\phi(z\mid c)$ (i.e., $Z\perp S\mid C$).
Denote the encoder-induced joint distribution by
\begin{equation}
 q(s,c,z)=p_{\text{env}}(s)\,q_\psi(c\mid s)\,q_\phi(z\mid c),
 \quad
 q(s,z)=\int q(s,c,z)\,dc .
\end{equation}

\subsection{Connecting Reconstruction to Mutual Information}

Our goal is to encourage the concepts $c$ to preserve information about the global state $s$, i.e., to increase $I(S;C)$.
We first show that reconstruction maximizes a variational lower bound of $I(S;Z)$.

\textbf{Proposition 1 (Variational information maximization).}
Maximizing the expected log-likelihood $\mathbb E_{q(s,z)}[\log p_\theta(s\mid z)]$ maximizes a variational lower bound of $I_q(S;Z)$.

\textit{Proof.}
By definition,
\begin{equation}
  I_q(S;Z)=H_q(S)-H_q(S\mid Z)=H_q(S)+\mathbb E_{q(s,z)}[\log q(s\mid z)].
\end{equation}
Introduce a variational decoder $p_\theta(s\mid z)$. Using the non-negativity of KL divergence,
\begin{equation}
 \mathbb E_{q(s\mid z)}[\log q(s\mid z)] \ge \mathbb E_{q(s\mid z)}[\log p_\theta(s\mid z)],
\end{equation}
we obtain the Barber--Agakov lower bound:
\begin{equation}
 \label{eq:variational_bound}
 I_q(S;Z) \ge H_q(S) + \mathbb E_{q(s,z)}[\log p_\theta(s\mid z)].
\end{equation}
Since $H_q(S)$ is constant w.r.t.\ $(\phi,\theta)$, maximizing $\mathbb E_{q(s,z)}[\log p_\theta(s\mid z)]$
(or minimizing the negative log-likelihood reconstruction loss) maximizes this lower bound. \hfill $\square$

\subsection{Enforcing Concept Sufficiency via DPI}

\textbf{Proposition 2 (Sufficiency propagation via DPI).}
Increasing $I_q(S;Z)$ pushes up a certified lower bound of $I_q(S;C)$.

\textit{Proof.}
From the Markov chain $S\rightarrow C\rightarrow Z$, the Data Processing Inequality (DPI) implies
\begin{equation}
    \label{eq:dpi}
    I_q(S;C)\ge I_q(S;Z).
\end{equation}
Combining Eq.~(\ref{eq:variational_bound}) and Eq.~(\ref{eq:dpi}) yields
\begin{equation}
    I_q(S;C)\;\ge\; I_q(S;Z)\;\ge\; H_q(S)+\mathbb E_{q(s,z)}[\log p_\theta(s\mid z)].
\end{equation}
Therefore, improving reconstruction from $z$ increases a guaranteed lower bound on $I_q(S;C)$:
since $Z$ is computed solely from $C$, $Z$ cannot contain information about $S$ that is not already present in $C$.
Thus, demanding accurate reconstruction $Z\to S$ forces the upstream concept representation $C$ to preserve globally predictive information. \hfill $\square$

\subsection{Role of the Bottleneck and KL Regularization (Variational IB)}

We optimize a bottlenecked objective of the form
\begin{equation}
     \mathcal L_{\text{aux}}
     = \mathbb E_{q(s,c)}\mathbb E_{q_\phi(z\mid c)}[\log p_\theta(s\mid z)]
     - \beta\,\mathbb E_{q(c)}\mathrm{KL}\!\big(q_\phi(z\mid c)\,\|\,p(z)\big),
\end{equation}
where $p(z)$ is a simple prior. The KL term imposes an information bottleneck on the channel $C\to Z$.
In particular, under $q(c)q_\phi(z\mid c)$, we have the decomposition
\begin{equation}
     \mathbb E_{q(c)}\mathrm{KL}\!\big(q_\phi(z\mid c)\,\|\,p(z)\big)
    = I_q(C;Z)+\mathrm{KL}\!\big(q(z)\,\|\,p(z)\big),
\end{equation}
showing that the regularizer controls the information capacity $I_q(C;Z)$ and aligns the aggregated posterior $q(z)$ with the prior.
Hence, the auxiliary objective implements a Variational Information Bottleneck for predicting $S$ from $C$:
it encourages $Z$ (and thus $C$) to retain compressible, robust information that is maximally predictive of the global state.

\section{Pseudo Code}\label{pseudocode}
\begin{algorithm}[H]
   \caption{Escaping Local Views (ELV)}
   \label{alg:elv}
\begin{algorithmic}
   \STATE Initialize a set of agents $\mathcal{N} = \{1, \dots, n\}$
   \STATE Initialize concept networks and policy networks $Q_i(\tau_i, u_i; \theta)$, target networks parameterized by $\hat{\theta} \leftarrow \theta$
   \STATE Initialize VAE components, concept predictor and concept decoder with parameters $\phi$
   \STATE Initialize a replay buffer $\mathcal{B}$ for storing episodes
   \REPEAT
      \STATE Initialize a history state $h_i^0$ and previous action $u_i^0$ for each agent
      \STATE Observe each agent's partial observation $[o_i^1]_{i=1}^n$
      \FOR{$t = 1 : T$}
         \STATE Get current history $h_i^t$ and extract latent concepts $c_i^t$
         \STATE Infer global latent context $z$ from concepts and calculate value function $Q_i(\tau_i^t, u_i^{t})$
         \STATE Select action $u_i^t$ via $\epsilon$-greedy exploration based on $Q_i$
         \STATE Execute joint action $\boldsymbol{u}^t$ to receive reward $r^t$ and next state $s^{t+1}$
      \ENDFOR
      \STATE Store the episode trajectory to $\mathcal{B}$
      \STATE Sample a batch of episode trajectories with batch size $b$ from $\mathcal{B}$
      \FOR{$t = 1 : T$}
         \STATE Predict next concepts $\hat{c}^{t+1}$ and calculate intrinsic reward $r^{int}_t$ via prediction error in Eq.~(\ref{F:intrinsic reward})
         \STATE Calculate augmented reward $r_t = r^{env}_t + \beta_t \cdot r^{int}_t$
         \STATE Get reconstructed observation $\tilde{o}_i^t$ from concepts and calculate grounding loss $\mathcal{L}_{rec}$ via Eq.~(\ref{F:recloss})
         \STATE Construct global latent variable $z \sim q_\phi(z|c)$ and calculate VAE loss $\mathcal{L}_{vae}$ via Eq.~(\ref{F:vaeloss})
         \STATE Calculate concept importance weights $\alpha_k, \beta_k$ using dual-path attention in Eq.~(\ref{F:importanceweight}) and~(\ref{eq:second_attn})
         \STATE Compute individual Q-values incorporating order-2 concept interactions via Eq.~(\ref{Eq:computeQ})
      \ENDFOR
      \STATE Construct the total loss function defined in Eq.~(\ref{eq:total_loss})
      \STATE Update $\theta$ and $\phi$ by minimizing the total loss
      \STATE Periodically update target parameters $\hat{\theta} \leftarrow \theta$
   \UNTIL{$Q_i$ converges}
\end{algorithmic}
\end{algorithm}

\section{Related Work}\label{RW}
\textbf{Interpretability in MARL.}
The opacity of deep MARL policies has traditionally been addressed via post-hoc explanations, such as saliency-based analyses~\citep{greydanus2018visualizing} or Shapley-style attributions for value decomposition~\citep{wang2020shapley}.
While informative, post-hoc explanations can be sensitive to perturbations and may provide primarily correlational evidence rather than a structured account of multi-agent coordination~\citep{slack2021reliable}.
This motivates intrinsically interpretable architectures that expose intermediate decision structure, e.g., differentiable trees or symbolic distillation~\citep{bastani2018verifiable,liu2025mixrts}.
However, such designs may struggle to simultaneously maintain readability and expressiveness when coordination is high-dimensional and nontrivial. Graph-based interpretability methods can reveal who interacts~\citep{wu2025gcm}, but may not directly ground what semantic content is exchanged.
In contrast, ELV emphasizes cross-agent semantic coherence by training concepts to be globally informative under a VAE-style reconstruction objective, mitigating fragmented local views.

\textbf{Concept Bottleneck Models (CBMs).}
Concept Bottleneck Models (CBMs) introduce an interpretable intermediate layer where input features are first mapped to human-understandable concepts before making final predictions \citep{koh2020concept}. While early CBMs suffered from a ``performance-interpretability trade-off," Concept Embedding Models (CEMs) \citep{espinosa2022concept} mitigated this by representing concepts as high-dimensional vectors. To bypass the need for exhaustive manual annotations, Post-hoc CBMs \citep{yuksekgonulpost} and Label-free CBMs \citep{oikarinen2023label} leveraged pre-trained vision-language models like CLIP to automatically define concept spaces. Recent advances, such as Probabilistic CBMs \citep{kim2023probabilistic} and Soft-CBMs \citep{havasi2022addressing}, further enhance robustness by modeling concept uncertainty, ensuring that semantic abstractions remain reliable even under distribution shifts or noisy inputs.
Recent research has sought to bridge such semantic abstractions with reinforcement learning.
For instance, SCoBots~\citep{delfosse2024interpretable} successfully utilizes hierarchical concepts to align single-agent policies, yet extending this transparency to decentralized coordination remains non-trivial.

\textbf{Semantic Exploration and Intrinsic Motivation.}
Efficient exploration in sparse-reward MARL remains challenging.
Classic intrinsic motivation methods such as curiosity or RND~\citep{pathak2017curiosity,burda2018exploration} often operate in raw observation spaces and can be distracted by task-irrelevant stochasticity.
Beyond this, recent studies observe that na\"ively mixing intrinsic and extrinsic objectives may bias optimization and cause ``over-exploration'';
accordingly, constrained or adaptive schemes have been proposed to automatically regulate intrinsic reward strength and reduce such interference~\citep{chen2022}.
To further improve robustness, uncertainty-aware curiosity explicitly down-weights inherently unpredictable (aleatoric) transitions, mitigating the noisy-TV failure mode~\citep{mavor2022stay}.
Meanwhile, structure-aware intrinsic rewards leverage transition regularities (e.g., successor--predecessor representations) to encourage bottleneck-seeking exploration beyond purely local novelty~\citep{yu2023successor}.
Recent methods also incorporate social context or calibration to stabilize intrinsic signals~\citep{pan2025wonder}.
ELV is complementary: it computes prediction errors in a learned latent concept space, encouraging exploration towards semantically uncertain transitions rather than pixel-level noise~\citep{mccaffrey2025predictive}.

\begin{figure}[ht]
\centering
\subfloat[Level Based Foraging]{
    \paperfigure{0.22\linewidth}{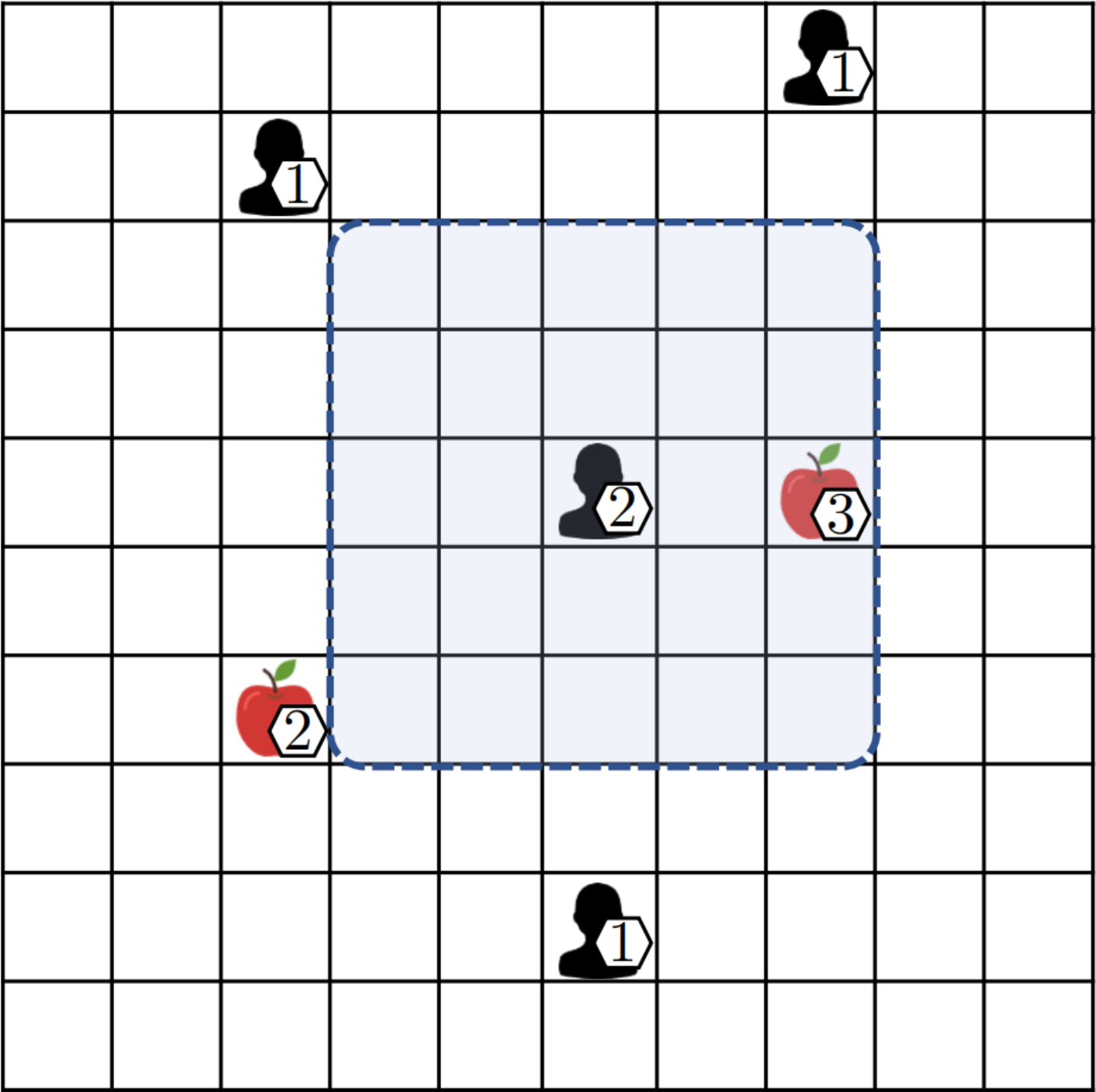}
}
\subfloat[StarCraft Multi-Agent Challenge]{
    \paperfigure{0.33\linewidth}{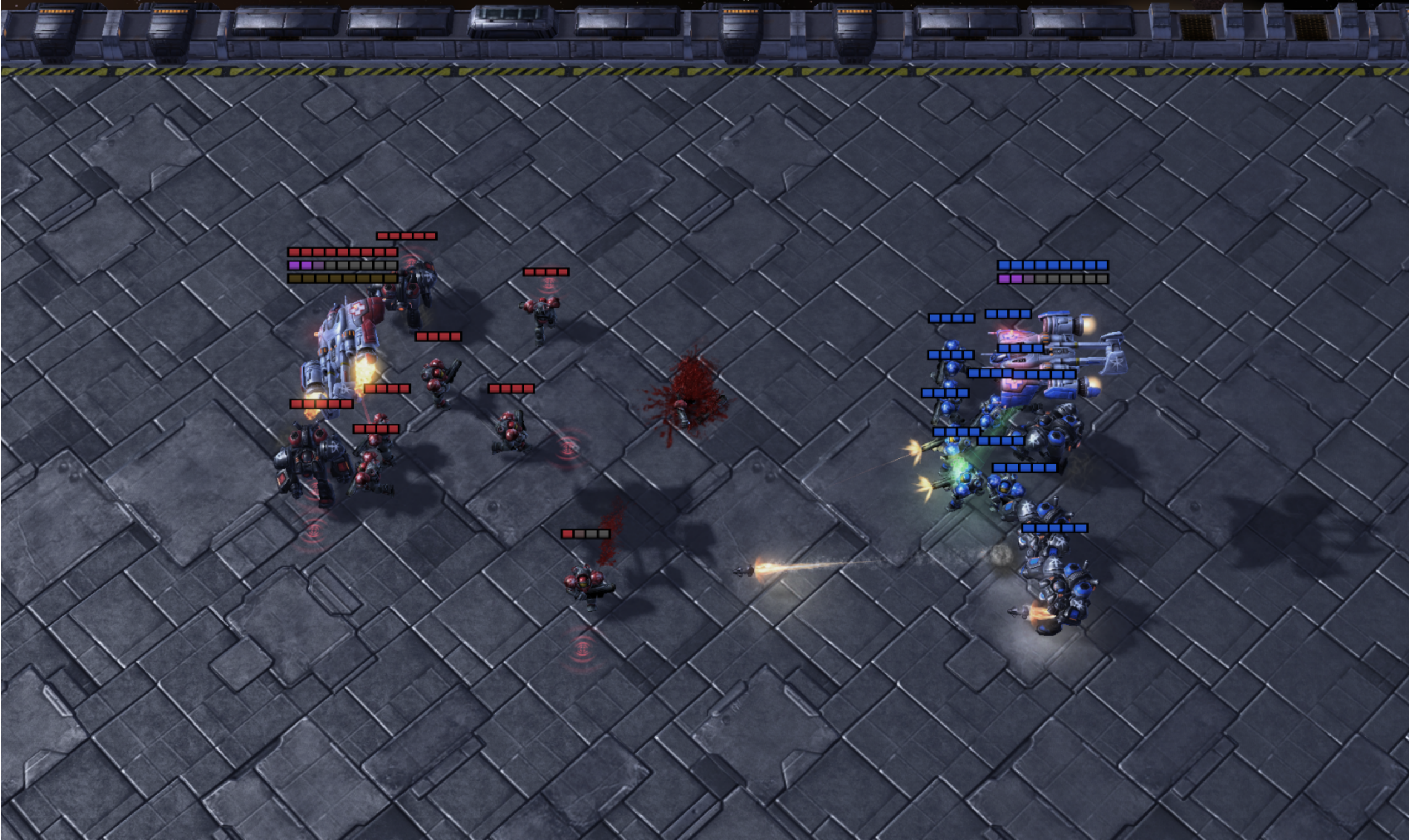}
}
\subfloat[StarCraft Multi-Agent Challenge v2]{
    \paperfigure{0.35\linewidth}{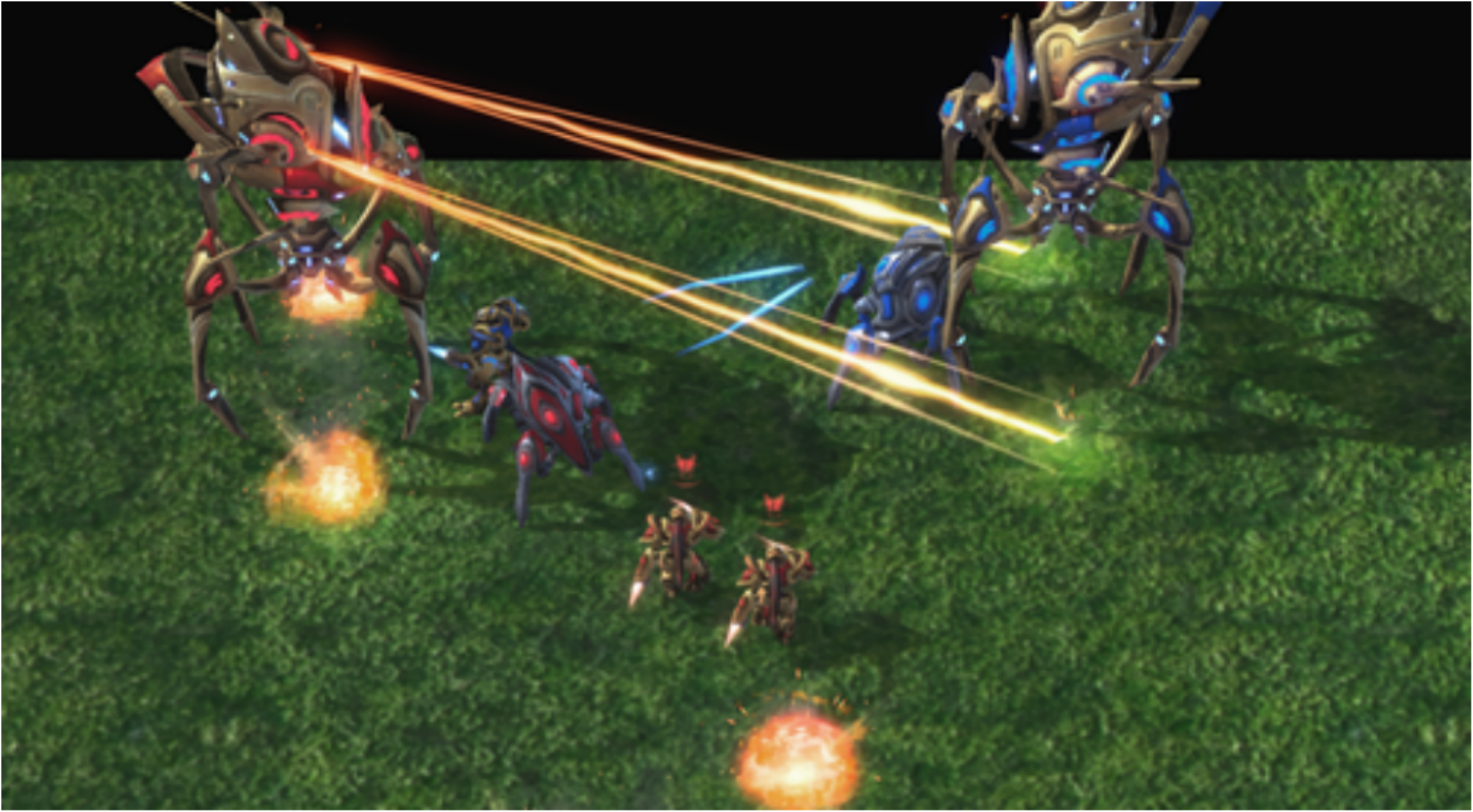}
}
\caption{Three benchmarks used in our experiments.}
\label{fig:benchmark_demo}
\end{figure}

\section{Experimental Details}\label{ExperimentalDetails}

\subsection{Benchmark and Settings}

The majority of our experiments are carried out on Level Based Foraging (LBF), StarCraft Multi-Agent Challenge (SMAC) and SMACv2, as shown in Fig.~\ref{fig:benchmark_demo}. We describe the details and settings of these benchmarks in this section. Comprehensive experimental details are enumerated in Table~\ref{tab:lbf_settings} and Table~\ref{tab:smac_settings}.

\begin{table}[htbp]
\caption{Experimental Settings of Level-Based Foraging.}
\label{tab:lbf_settings}
\centering
\begin{small} 
\begin{tabularx}{\linewidth}{@{}l l >{\raggedright\arraybackslash}X@{}}
\toprule
\textsc{Hyperparameter} & \textsc{Value} & \textsc{Description} \\
\midrule
\textsc{Max Player Level} & 3 & \textsc{Maximum Agent Level Attribute} \\
\textsc{Max Episode Length} & 50 & \textsc{Maximum Timesteps Per Episode} \\
\textsc{Batch Size} & 32 & \textsc{Number of Episodes Per Update} \\
\textsc{Test Interval} & 10,000 & \textsc{Frequency of Evaluating Performance} \\
\textsc{Test Episodes} & 32 & \textsc{Number of Episodes to Test} \\
\textsc{Replay Batch Size} & 5000 & \textsc{Maximum Number of Episodes Stored in Memory} \\
\textsc{Discount Factor} $\gamma$ & 0.99 & \textsc{Degree of Impact of Future Rewards} \\
\textsc{Total Timesteps} & 1,050,000 & \textsc{Number of Training Steps} \\
\textsc{Start} $\epsilon$ & 1.0 & \textsc{The Start $\epsilon$ Value to Explore} \\
\textsc{Finish} $\epsilon$ & 0.05 & \textsc{The Finish $\epsilon$ Value to Explore} \\
\textsc{Anneal Steps for} $\epsilon$ & 50,000 & \textsc{Number of Steps of Linear Annealing} \\
\textsc{Target Update Interval} & 200 & \textsc{The Target Network Update Cycle} \\
\bottomrule
\end{tabularx}
\end{small}
\end{table}

\begin{table}[htbp]
\caption{Experimental settings of SMAC and SMACv2.}
\label{tab:smac_settings}
\centering
\begin{small} 
\begin{tabularx}{\linewidth}{@{}l l >{\raggedright\arraybackslash}X@{}}
\toprule
\textsc{Hyperparameter} & \textsc{Value} & \textsc{Description} \\
\midrule
\textsc{Difficulty} & 7 & \textsc{Enemy Units with Built-in AI Difficulty} \\
\textsc{Batch Size (SMAC)} & 32 & \textsc{Number of Episodes Per Update} \\
\textsc{Batch Size (SMACv2)} & 128 & \textsc{Number of Episodes Per Update} \\
\textsc{Test Interval} & 10,000 & \textsc{Frequency of Evaluating Performance} \\
\textsc{Test Episodes} & 32 & \textsc{Number of Episodes to Test} \\
\textsc{Replay Batch Size} & 5000 & \textsc{Maximum Number of Episodes in Memory} \\
\textsc{Discount Factor} $\gamma$ & 0.99 & \textsc{Degree of Impact of Future Rewards} \\
\textsc{Start} $\epsilon$ & 1.0 & \textsc{The Start $\epsilon$ Value to Explore} \\
\textsc{Finish} $\epsilon$ & 0.05 & \textsc{The Finish $\epsilon$ Value to Explore} \\
\textsc{Anneal Steps for Easy \& Hard} & 50,000 & \textsc{Number of Steps of Linear Annealing $\epsilon$} \\
\textsc{Anneal Steps for Super Hard} & 100,000 & \textsc{Number of Steps of Linear Annealing $\epsilon$} \\
\textsc{Target Update Interval} & 200 & \textsc{The Target Network Update Cycle} \\
\bottomrule
\end{tabularx}
\end{small}
\end{table}

\begin{table}[t] 
\centering
\caption{The detailed specifications of SMAC scenarios used in our evaluation.}
\label{tab:smac_configs}

\resizebox{\textwidth}{!}{
\begin{tabular}{lllll}
\toprule
\textsc{Map Name} & \textsc{Ally Units} & \textsc{Enemy Units} & \textsc{Timesteps} & \textsc{Scenario Type} \\
\midrule
2s3z & 2 \textsc{Stalkers} \& 3 \textsc{Zealots} & 2 \textsc{Stalkers} \& 3 \textsc{Zealots} & 2M & \textsc{Easy} \\
3s5z & 3 \textsc{Stalkers} \& 5 \textsc{Zealots} & 3 \textsc{Stalkers} \& 5 \textsc{Zealots} & 2M & \textsc{Easy} \\
1c3s5z & 1 \textsc{Colossus}, 3 \textsc{Stalkers} \& 5 \textsc{Zealots} & 1 \textsc{Colossus}, 3 \textsc{Stalkers} \& 5 \textsc{Zealots} & 2M & \textsc{Easy} \\
2s\_vs\_1sc & 2 \textsc{Stalkers} & 1 \textsc{Spine Crawler} & 2M & \textsc{Easy} \\
\midrule
5m\_vs\_6m & 5 \textsc{Marines} & 6 \textsc{Marines} & 2M & \textsc{Hard} \\
8m\_vs\_9m & 8 \textsc{Marines} & 9 \textsc{Marines} & 2M & \textsc{Hard} \\
3s\_vs\_5z & 3 \textsc{Stalkers} & 5 \textsc{Zealots} & 2M & \textsc{Hard} \\
2c\_vs\_64zg & 2 \textsc{Colossi} & 64 \textsc{Zerglings} & 2M & \textsc{Hard} \\
\midrule
MMM2 & 1 \textsc{Medivac}, 2 \textsc{Marauders} \& 7 \textsc{Marines} & 1 \textsc{Medivac}, 3 \textsc{Marauders} \& 8 \textsc{Marines} & 2M & \textsc{Super Hard} \\
3s5z\_vs\_3s6z & 3 \textsc{Stalkers} \& 5 \textsc{Zealots} & 3 \textsc{Stalkers} \& 6 \textsc{Zealots} & 5M & \textsc{Super Hard} \\
6h\_vs\_8z & 6 \textsc{Hydralisks} & 8 \textsc{Zealots} & 5M & \textsc{Super Hard} \\
corridor & 6 \textsc{Zealots} & 24 \textsc{Zerglings} & 5M & \textsc{Super Hard} \\
\bottomrule
\end{tabular}
}
\end{table}

\begin{table}[t]
\caption{The SMACv2 scenarios used in our evaluation.}
\label{tab:smacv2_config}
\centering
\begin{small}
\begin{tabularx}{\linewidth}{@{}lccc>{\raggedright\arraybackslash}X@{}}
\toprule
\textsc{Scenario Name} & \textsc{Allies} & \textsc{Enemies} & \textsc{Timesteps} & \textsc{Unit Composition (Randomized)} \\
\midrule
Terran\_5\_vs\_5 & 5 & 5 & 1M & \multirow{2}{=}{\textsc{Marines, Marauders, Medivacs}} \\
Terran\_10\_vs\_10 & 10 & 10 & 2M & \\
\midrule
Zerg\_5\_vs\_5 & 5 & 5 & 1M & \multirow{2}{=}{\textsc{Zerglings, Hydralisks, Banelings}} \\
Zerg\_10\_vs\_10 & 10 & 10 & 2M & \\
\midrule
Protoss\_5\_vs\_5 & 5 & 5 & 1M & \multirow{2}{=}{\textsc{Stalkers, Zealots, Colossi}} \\
Protoss\_10\_vs\_10 & 10 & 10 & 2M & \\
\bottomrule
\end{tabularx}
\end{small}
\end{table}

\textbf{Level-Based Foraging (LBF).} LBF simulates a cooperative multi-agent task on a $10 \times 10$ grid, where agents with initialized skill levels must coordinate to collect food items under partial observability (restricted to a $5 \times 5$ local view). The core challenge lies in its enforced cooperation mechanism: a food item with difficulty $L_f$ can only be consumed if the sum of the levels of participating agents meets the threshold ($\sum l_i \ge L_f$), necessitating dynamic subgroup formation. To encourage efficient exploration, the environment applies a step penalty of $-0.002$ at each timestep, and we evaluate our method on two specific configurations: 4-Players \& 2-Foods and 6-Players \& 4-Foods.

\textbf{StarCraft Multi-Agent Challenge.} The StarCraft Multi-Agent Challenge (SMAC)~\citep{SMAC} based on StarCraft II (version SC2.4.10), which provides a suite of cooperative multi-agent reinforcement learning (MARL) tasks where a team of allied units is trained to defeat enemy units controlled by the built-in StarCraft II AI. The enemy AI operates at difficulty level $7$ in all scenarios.Each allied unit is controlled by an individual decentralized agent, which observes the environment locally and learns to act cooperatively through reinforcement learning. Agents are trained to maximize cumulative rewards by dealing as much damage as possible to enemy units while avoiding unnecessary damage to themselves. These tasks demand a wide range of cooperative behaviors, including focus fire, kiting, flanking, and spatial coordination.
The evaluation is performed across multiple challenging combat scenarios with varying levels of complexity, agent types, and asymmetry. Some scenarios involve homogeneous units, while others include heterogeneous unit compositions. A summary of the selected scenarios is provided in Table ~\ref{tab:smac_configs}.

\textbf{StarCraft Multi-Agent Challenge v2.} StarCraft Multi-Agent Challenge v2 (SMACv2) \citep{ellis2023smacv2} is a significant evolution of the original benchmark designed to rigorously test the generalization capabilities of MARL algorithms. Unlike the static scenarios in v1, SMACv2 introduces a high degree of stochasticity through randomized initial configurations, where enemy positions, unit attributes, and quantities vary across episodes. Furthermore, the inclusion of dynamic environmental features—such as map alterations and fluctuating objectives—adds a layer of unpredictability. All experiments are performed using the SC2.4.10 environment with uniform training step limits to ensure a fair and robust assessment of scalability and adaptability. Table~\ref{tab:smacv2_config} summarizes the detailed specifications for each scenario.

\subsection{Hyperparameters of Baselines}

We implement a diverse set of baselines using the PyMARL framework~\citep{SMAC}. These include five classic value-decomposition methods---such as QMIX~\citep{rashid2020monotonic}, VDN~\citep{sunehag2018value}, QPLEX~\citep{wang2021qplex}, CDS~\citep{li2021celebrating} and QTRAN~\citep{son2019qtran}---as well as recent approaches focusing on exploration and communication, including $\mathrm{NA^2Q}$~\citep{liu2023na2q}, ReBorn~\citep{qin2024dormant}, $\mathrm{MA^2E}$~\citep{kangma}, and SHAQ~\citep{wang2022shaq}. For fair comparison, all baselines adhere to the default hyperparameters provided in their respective official implementations or the PyMARL codebase.

During evaluation, we employ decentralized greedy action selection, testing the policy every $20,000$ environment steps over $20$ episodes. To ensure statistical robustness, all experiments are conducted across five random seeds in environments characterized by clear success/failure dynamics. The metric \textit{test win rate} is defined as the percentage of episodes where allied agents successfully eliminate all enemies within the time limit. Optimization is performed using RMSprop with a learning rate of $5 \times 10^{-4}$, with target networks updated every $200$ training episodes.

\subsection{Hyperparameters of ELV}
We implement ELV based on the PyMARL framework and optimize the entire ELV architecture using the Adam optimizer with a learning rate of $5\times 10^{-4}$.
The target networks are updated every 200 training episodes to stabilize value estimation.
For a detailed listing of the structural hyperparameters and loss coefficients specific to ELV, please refer to Table~\ref{tab:elv_hyperparams}.

\begin{table}[htbp]
    \centering
    \caption{Hyper-parameters of ELV.}
    \label{tab:elv_hyperparams}
    \begin{tabular}{lcr} 
        \toprule
        \textsc{Component} & \textsc{Hyper-parameters} & \textsc{Value} \\
        \midrule
        & \textsc{Concept Number} ($K$) & 16 \\
        \textsc{Concept} & \textsc{Concept Dimension} ($d$) & 16 \\
        \textsc{Extraction} & \textsc{History Hidden Dim} & 64 \\
        \midrule
        & \textsc{Latent Context Dim} ($z$) & 32 \\
        \textsc{Variational} & \textsc{VAE Loss Weight} $\lambda_1$ & 0.1 \\
        \textsc{Autoencoder} & \textsc{Rec. Loss Weight} $\lambda_2$ & 0.1 \\
        \midrule
        & \textsc{Pred. Loss Weight} $\lambda_3$ & 0.05 \\
        \textsc{Intrinsic} & \textsc{Initial Scale} $\beta_0$ & 1 \\
        \textsc{Reward} & \textsc{Decay Coefficient} $\lambda$ & $9.2 \times 10^{-8}$ \\ 
        & \textsc{Decay Schedule} & Exponential \\
        \midrule
        \textsc{Dual-Path} & \textsc{Query Dimension} & 16 \\
        \textsc{Concept Importance} & \textsc{Key Dimension} & 16 \\
        \bottomrule
    \end{tabular}
\end{table}

\newpage
\begin{figure}[!ht]
	\paperfigure{\textwidth}{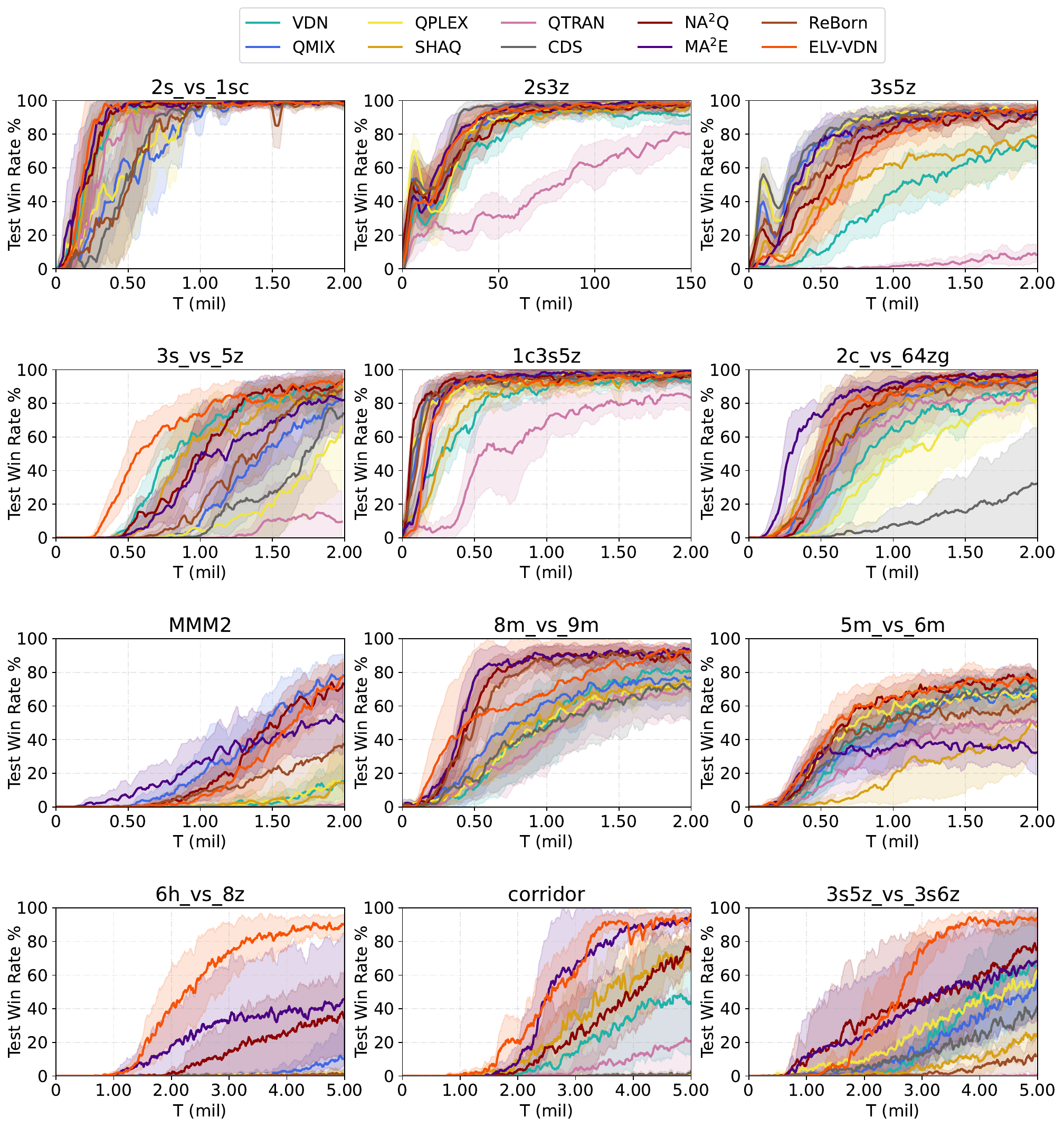}
	\caption{Median test win rate \% on ELV with VDN mixing network.}
	\label{ELV_vdn}
\end{figure}
\section{Additional Experimental Results}\label{AdditionalExperimental}

\begin{figure}[ht]
	\paperfigure{\textwidth}{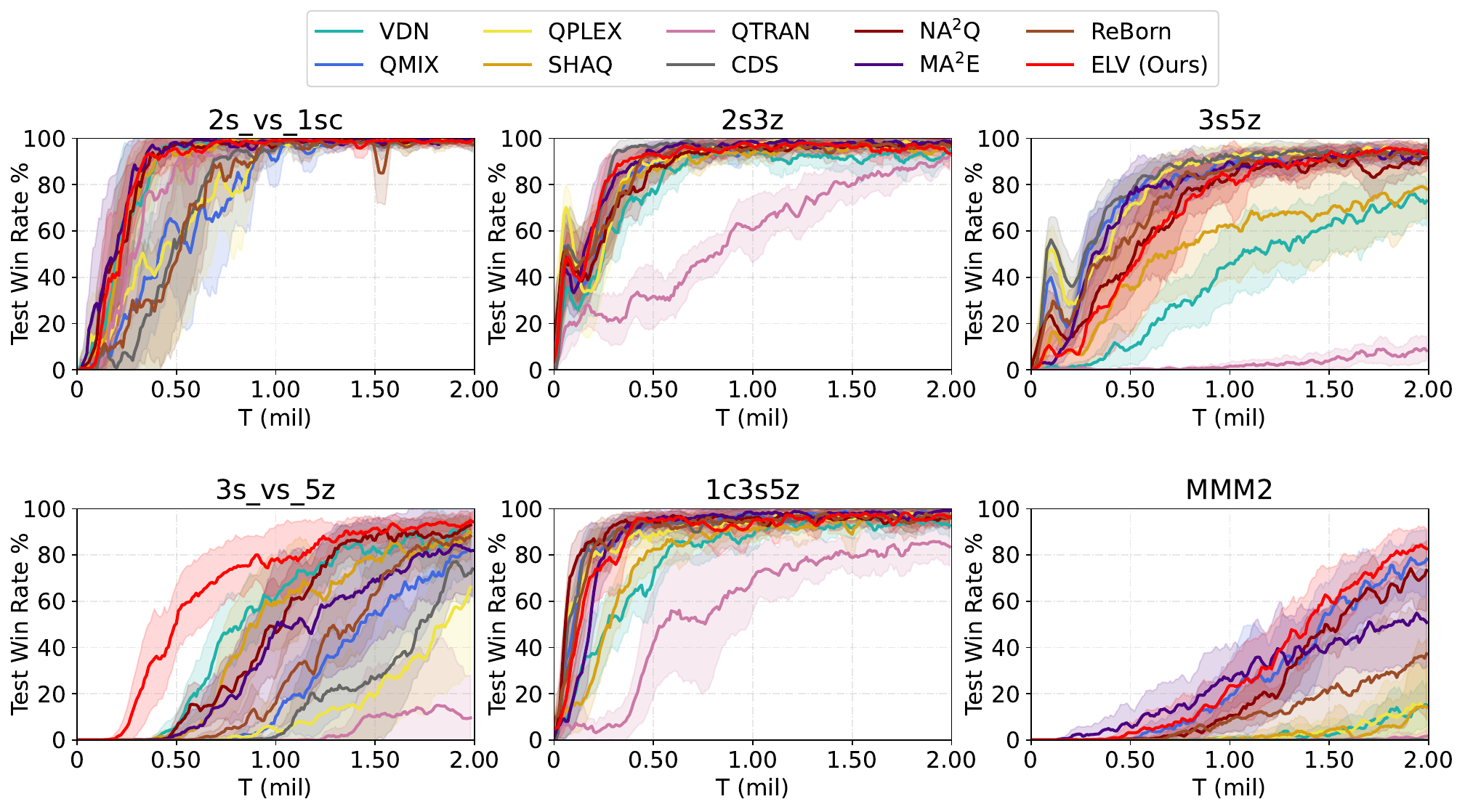}
	\caption{Test win rate \% for six extra scenarios of SMAC benchmark.}
	\label{easy_smac}
\end{figure}

\begin{figure}[!ht]
	\paperfigure{\textwidth}{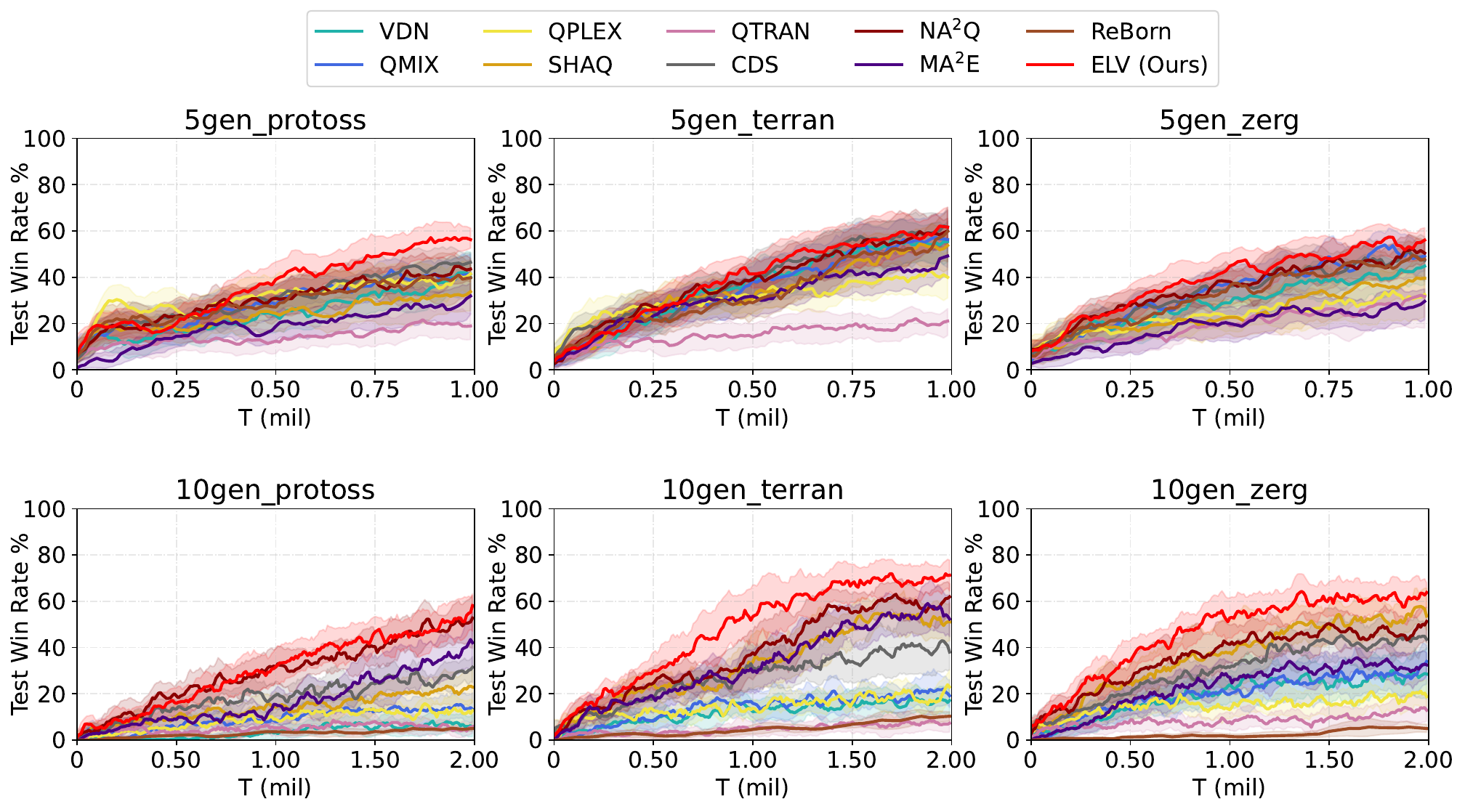}
	\caption{Median test win rate \% on SMACv2.}
	\label{smacv2}
\end{figure}

\subsection{Generality across Mixing Architectures.}\label{elvvdn}
Since the proposed ELV framework focuses on restructuring the local agent's policy network via latent concept extraction, it is theoretically compatible with arbitrary global mixing networks. To empirically validate this architectural flexibility, we integrated ELV with Value-Decomposition Networks (VDN), which employs a simple linear summation for value aggregation, contrasting with the non-linear mixing of QMIX used in our main experiments.
As illustrated in \textit{Figure}~\ref{ELV_vdn}, ELV-VDN demonstrates remarkable performance improvements over the vanilla VDN baseline. This advantage is particularly evident in super-hard scenarios such as 6h\_vs\_8z and corridor. In these tasks, the standard VDN fails to learn effective policies due to its limited representational capacity, hovering near a 0\% win rate. In stark contrast, ELV-VDN achieves high win rates competitive with distinctively more complex architectures like QMIX and QPLEX. This result underscores that the semantic concepts learned by ELV provide a robust inductive bias, enabling simple mixing mechanisms to solve complex coordination problems. It confirms that ELV is a generalizable framework compatible with diverse value factorization backbones.

\subsection{Additional Experimental Results on SMAC Extra Scenarios.}\label{extrasmacqmix}
To verify the stability of ELV across varying difficulty levels, we further evaluate its performance on six easy SMAC scenarios. As illustrated in \textit{Figure}~\ref{easy_smac}, most baselines, such as QMIX and QPLEX, achieve near-saturated win rates in these environments due to the lower coordination threshold.
ELV consistently matches this optimal performance, converging rapidly to a 100\% win rate. This result is crucial as it confirms that the introduction of the latent concept bottleneck and the auxiliary reconstruction objectives does not induce performance regression or instability in simpler tasks. The learned semantic concepts efficiently collapse into simple heuristic policies when complex coordination is unnecessary, demonstrating the robustness of the ELV framework.

\subsection{Performance on SMACv2.}\label{PerformanceonSMACv2}
Performance on SMACv2. We further assess the generalization capability of ELV under the high stochasticity of SMACv2, where unit compositions and start positions are randomized. As shown in \textit{Figure}~\ref{smacv2}, ELV consistently achieves superior win rates across all scenarios, establishing a significant performance margin in the more complex 10-unit tasks (e.g., 10gen\_terran and 10gen\_zerg).
Notably, CDS, despite being designed for heterogeneous agents, exhibits high variance and suboptimal convergence. This failure suggests that merely encouraging behavioral diversity is insufficient to adapt to dynamic tactical shifts without grounded semantic understanding. Similarly, ReBorn struggles in large-scale settings (10gen), implying that its exploration mechanism may be inefficient in handling the exponentially growing state space of randomized scenarios. In contrast, QMIX and QPLEX require significantly more samples to reach competitive performance, struggling with the non-stationarity of the environment. ELV's success stems from its ability to extract invariant semantic concepts from variable observations, effectively filtering out stochastic noise and enabling robust coordination where baselines falter.

\begin{figure}[t]
	\paperfigure{\textwidth}{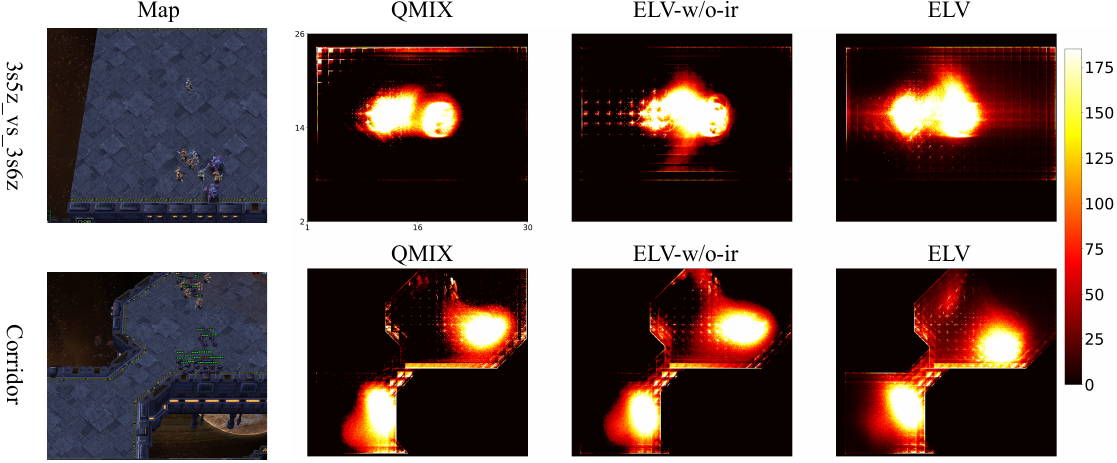}
	\caption{Visualization of exploration coverage via spatial heatmaps.}
	\label{ELV_exp}
\end{figure}

\subsection{Qualitative Analysis of Exploration.}

To intuitively verify how the proposed concept-based intrinsic reward facilitates exploration, we visualize the spatial heatmaps of agent trajectories throughout the training process in \textit{Figure}~\ref{ELV_exp}. In the super-hard scenario 3s5z\_vs\_3s6z, which demands complex kiting and positioning, QMIX exhibits a highly concentrated visitation pattern, indicating that agents tend to stagnate in local optima with limited movement. Although removing the intrinsic reward (ELV-w/o-ir) slightly expands the range, it still fails to cover the map effectively. In contrast, the full ELV model demonstrates significantly broader spatial coverage, suggesting that agents actively utilize the entire terrain to execute superior micro-tactics.
In the navigation-heavy Corridor map (Bottom Row), the disparity is even more pronounced. Agents must pass through a narrow bottleneck to reach enemies. Baselines like QMIX and ELV-w/o-ir show high-density clustering at the bottom-left entry, failing to effectively penetrate the choke point. ELV, however, not only successfully traverses the narrow passage but also achieves a more even and widespread distribution within the top-right target area relative to the baselines. This indicates that the intrinsic reward effectively drives agents to break through bottlenecks and explore the task-critical region more thoroughly, rather than clustering at the entrance.

\clearpage
\section{Additional Analysis of Concepts and Partial Observability}
\label{app:concept_analysis}

We further examine the learned concept representations from three complementary perspectives: their semantic roles and influence on policy behavior, their evolution along an episode, and their effectiveness under different degrees of partial observability.

\subsection{Semantic Roles and Test-Time Concept Interventions}
\label{app:concept_interventions}

\textbf{Semantic roles of learned concepts.}
ELV learns concept representations without ground-truth concept annotations.
To interpret the resulting slots, we use the concept--feature attribution matrix $G$ from Section~\ref{Explanation}, which measures the sensitivity of each observation modality's reconstruction to each concept.
The maximum-attribution matching yields a semantic assignment of concept slots to structured observation features.
Table~\ref{tab:concept_semantics} presents the resulting assignments for a representative model on \textit{3s\_vs\_5z}.
The assigned concepts cover movement, individual enemies, teammates, self-state, unit type, and action history, providing a concrete interpretation of the concept--feature structure visualized in Figure~\ref{concept assignment}.
For example, C2 is associated with tracking enemy unit 1, while C11 and C14 are associated with self-state and movement direction, respectively.

\begin{table}[!htbp]
    \centering
    \caption{Attribution-based semantic assignments of learned concept slots on \textit{3s\_vs\_5z}. Concept IDs refer to slots in the analyzed model.}
    \label{tab:concept_semantics}
    \begin{tabular}{lll}
        \toprule
        Concept ID & Dominant feature & Semantic label \\
        \midrule
        C14 & move & Movement direction \\
        C2 & enemy1 & Enemy unit 1 tracking \\
        C13 & enemy2 & Enemy unit 2 tracking \\
        C9 & enemy3 & Enemy unit 3 tracking \\
        C5 & enemy4 & Enemy unit 4 tracking \\
        C0 & enemy5 & Enemy unit 5 tracking \\
        C15 & ally1 & Teammate 1 monitoring \\
        C6 & ally2 & Teammate 2 monitoring \\
        C11 & self & Self-state \\
        C12 & type & Unit type recognition \\
        C4 & last\_action & Action history \\
        \bottomrule
    \end{tabular}
\end{table}

\textbf{Influence on policy behavior.}
We next perform test-time interventions on a trained ELV model on \textit{3s\_vs\_5z} by setting the existence probability of one concept slot to either $0$ or $1$ during evaluation.
Under the concept mixture defined in Section~\ref{sec:concept-predictor}, these settings select the slot's absence or presence embedding, respectively.
Each intervention setting is evaluated over 10 episodes; the unmodified policy achieves a win rate of $1.00$.
Table~\ref{tab:concept_interventions} reports the results for all 11 assigned slots.

The effects differ substantially across concepts.
Setting the existence probability of C9 or C11 to $0$ reduces the win rate to $0.00$, while the same intervention on C14 or C2 reduces it to $0.10$.
In contrast, interventions on C15 and C6 have smaller effects, with win rates remaining between $0.80$ and $0.90$.
Fixing a concept's probability to $1$ also changes performance: for example, C2 yields a win rate of $0.20$, compared with $0.90$ for C4 and C6.
These results show that the concept variables materially influence policy behavior and contribute unequally to coordination.
The reductions under both intervention settings further highlight the importance of context-dependent concept gating rather than keeping a slot permanently present or absent.

\begin{table}[!htbp]
    \centering
    \caption{Test-time concept interventions on \textit{3s\_vs\_5z}. Each setting uses 10 evaluation episodes. The unmodified policy's win rate is $1.00$; all values are proportions.}
    \label{tab:concept_interventions}
    \begin{tabular}{ccc}
        \toprule
        Concept ID & Win rate ($p=0$) & Win rate ($p=1$) \\
        \midrule
        C9 & 0.00 & 0.30 \\
        C11 & 0.00 & 0.60 \\
        C14 & 0.10 & 0.70 \\
        C2 & 0.10 & 0.20 \\
        C13 & 0.60 & 0.60 \\
        C4 & 0.60 & 0.90 \\
        C5 & 0.60 & 0.70 \\
        C0 & 0.70 & 0.80 \\
        C12 & 0.80 & 0.70 \\
        C15 & 0.90 & 0.80 \\
        C6 & 0.90 & 0.90 \\
        \bottomrule
    \end{tabular}
\end{table}

\subsection{Temporal Evolution of a Semantically Assigned Concept}
\label{app:concept_temporal}

\begin{figure}[!htbp]
    \centering
    \paperfigure{0.9\textwidth}{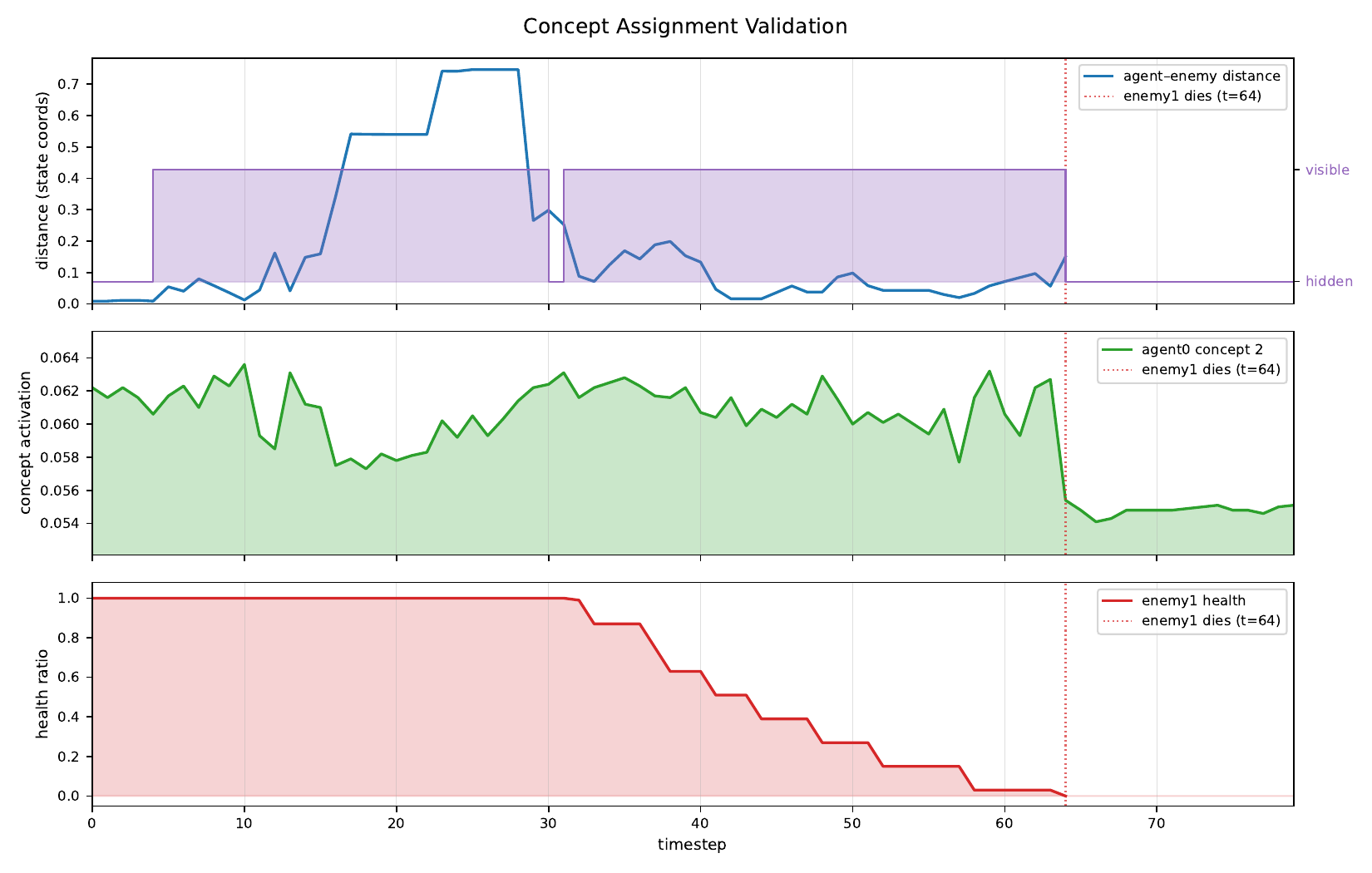}
    \caption{Temporal validation of the assignment of Agent 0's Concept 2 to enemy unit 1. The panels show agent--enemy distance with enemy visibility (top), concept activation (middle), and enemy health (bottom). The dotted vertical line marks the enemy death event at $t=64$.}
    \label{fig:concept_assignment_validation}
\end{figure}

To examine how an assigned concept evolves with its corresponding semantic factor, we follow Agent 0's Concept 2 together with enemy unit 1 over a full episode.
Figure~\ref{fig:concept_assignment_validation} shows the agent--enemy distance and enemy visibility, concept activation, and enemy health on a shared time axis.
The activation decreases as the agent--enemy distance grows during approximately $t=15$--$30$, then returns to a higher level when the enemy is nearby during approximately $t=30$--$63$.
At the marked enemy death event ($t=64$), activation drops sharply and remains lower for the rest of the episode.
This temporal pattern connects C2's attribution-based assignment to changes in the tracked enemy's state.
Together with the semantic assignment and intervention results, it illustrates how learned concept slots provide a meaningful interface for inspecting policy behavior over time.

\subsection{Performance under Restricted Local Observations}
\label{app:observation_radius}

\begin{figure}[!htbp]
    \centering
    \paperfigure{\textwidth}{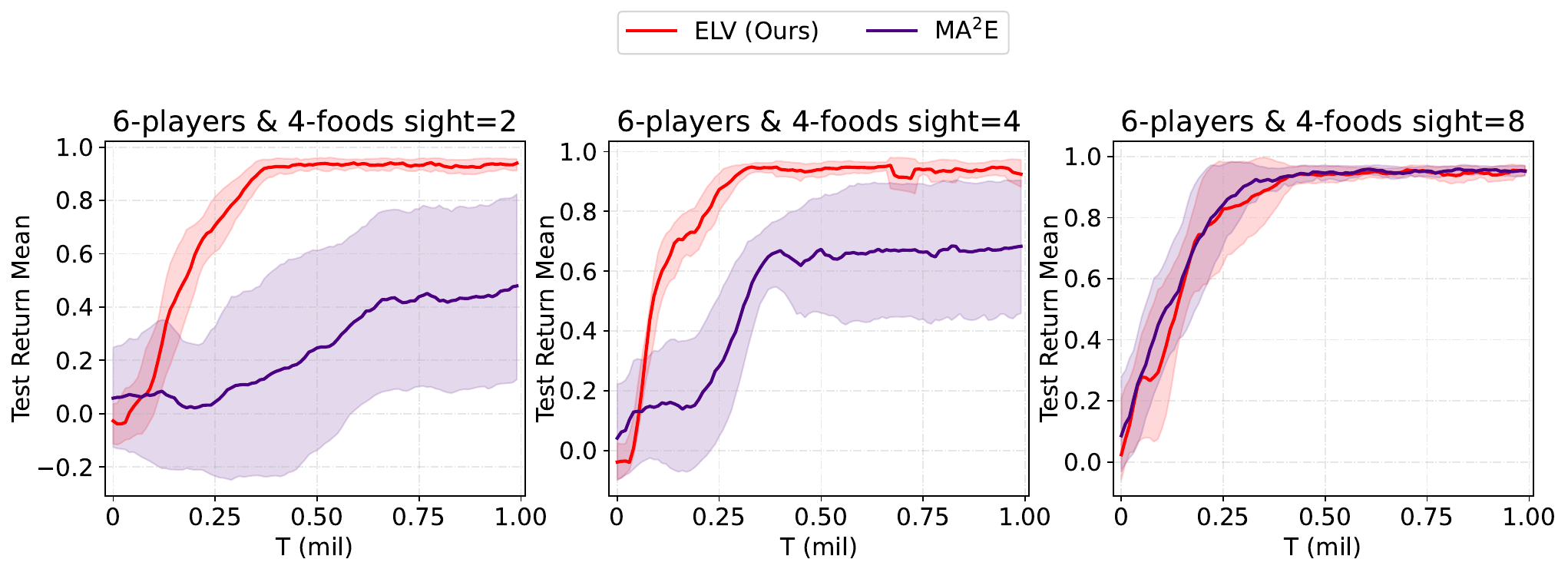}
    \caption{Mean test return on LBF with 6 players and 4 foods on a $10\times10$ grid, using observation radii of 2, 4, and 8. ELV and $\mathrm{MA^2E}$ use the same QMIX mixing network.}
    \label{fig:lbf_observation_radius}
\end{figure}

We evaluate the effect of partial observability in LBF by varying the local observation radius, $\mathrm{sight}\in\{2,4,8\}$, in the 6-players and 4-foods setting on a $10\times10$ grid.
Figure~\ref{fig:lbf_observation_radius} compares ELV with $\mathrm{MA^2E}$ over one million environment steps using the mean test return.
Both methods use the same QMIX mixing network, allowing the comparison to focus on their individual-value representations under restricted observations.

ELV maintains similar final returns across all three observation radii, including the most restrictive setting, $\mathrm{sight}=2$.
In contrast, $\mathrm{MA^2E}$ achieves substantially lower final returns at radii 2 and 4, while the two methods reach similar final performance at radius 8.
ELV also learns faster in the two more restricted settings.
The concentration of its advantage at smaller observation radii supports the role of the concept-based representation in handling limited local information, rather than attributing the improvement to a stronger global mixer.

\end{document}